%% file: main.tex
\documentclass[sigconf]{aamas}

\usepackage{balance} 

\doi{NGQQ5993}

\makeatletter
\gdef\@copyrightpermission{
  \begin{minipage}{0.2\columnwidth}
   \href{https://creativecommons.org/licenses/by/4.0/}{\includegraphics[width=0.90\textwidth]{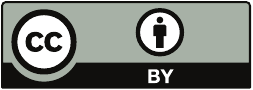}}
  \end{minipage}\hfill
  \begin{minipage}{0.8\columnwidth}
   \href{https://creativecommons.org/licenses/by/4.0/}{This work is licensed under a Creative Commons Attribution International 4.0 License.}
  \end{minipage}
  \vspace{5pt}
}
\makeatother

\setcopyright{ifaamas}
\acmConference[AAMAS '26]{Proc.\@ of the 25th International Conference
on Autonomous Agents and Multiagent Systems (AAMAS 2026)}{May 25 -- 29, 2026}
{Paphos, Cyprus}{C.~Amato, L.~Dennis, V.~Mascardi, J.~Thangarajah (eds.)}
\copyrightyear{2026}
\acmYear{2026}
\acmDOI{}
\acmPrice{}
\acmISBN{}

\input{sections/0a_preamble}

\input{sections/0b_acronyms}
\input{sections/0c_math_symbols}
\input{sections/0d_title}

\begin{abstract}
    \input{sections/1_abstract}
\end{abstract}


\begin{document}


\pagestyle{fancy}
\fancyhead{}


\maketitle 



\input{sections/2_introduction}
\input{sections/3_related_work}
\input{sections/4_problem_description} 
\input{sections/5_approach} 
\input{sections/6_evaluation}
\input{sections/7_conclusion}



\begin{acks}
This work was supported in part by the Army Research Lab under Grant W911NF-17-2-0181 (DCIST CRA).
We gratefully acknowledge Lae Young Kim for visualizing the motivating scenario.
\end{acks}



\bibliographystyle{ACM-Reference-Format} 
\bibliography{references}


\end{document}

%% file: sections/0a_preamble.tex
\usepackage{mathtools}
\usepackage{amsmath}
\usepackage{amsthm}
\usepackage{bm}

\usepackage{fancyref}
\usepackage{cleveref}

\usepackage{array}
\usepackage{tabularx}
\usepackage{multicol}
\usepackage{multirow}
\usepackage{booktabs}
\usepackage{bigstrut}  
\usepackage{arydshln}  

\usepackage{float}
\usepackage{stfloats}  
\usepackage{subcaption}
\usepackage{siunitx}

\usepackage{graphicx}
\usepackage[dvipsnames]{xcolor}
\usepackage{epsfig}
\usepackage[inkscapeexe=/Applications/Inkscape.app/Contents/MacOS/inkscape]{svg}

\usepackage{ifthen}
\usepackage{xparse}

\usepackage{pifont} 
\newcommand{\cmark}{\ding{51}} 
\newcommand{\xmark}{\ding{55}} 

\usepackage{romannum}

\graphicspath{ {./images/} }

\usepackage[ruled,vlined,linesnumbered]{algorithm2e}
\usepackage{listings}

\usepackage{glossaries}
\usepackage{enumitem}
\usepackage{tikz}

\newcommand{\cut}[1]{}

\newcommand{\figlabel}[1]{\label{fig:#1}}
\newcommand{\figref}[1]{Figure~\ref{fig:#1}}

\newcommand{\tablabel}[1]{\label{table:#1}}
\newcommand{\tabref}[1]{Table~\ref{table:#1}}

\newcommand{\equlabel}[1]{\label{equ:#1}}
\newcommand{\equref}[1]{Equation~(\ref{equ:#1})}
\newcommand{\equnum}[1]{(\ref{equ:#1})}

\newcommand{\constrref}[1]{Constraint~(\ref{equ:#1})}

\newcommand{\algolabel}[1]{\label{algo:#1}}
\newcommand{\algoref}[1]{Algorithm~\ref{algo:#1}}

\newcommand*\circled[1]{\tikz[baseline=(char.base)]{\node[shape=circle,draw,inner sep=0.7pt] (char) {\footnotesize{#1}};}}

\newcommand{\parens}[1]{\left(#1\right)}
\newcommand{\brackets}[1]{\left\{#1\right\}}
\newcommand{\angled}[1]{\left\langle#1\right\rangle}

\newcommand{\sqBrackets}[1]{\left[#1\right]}
\newcommand{\ooint}[1]{\parens{#1}}
\newcommand{\ccint}[1]{\sqBrackets{#1}}

\newcommand{\ocint}[1]{\left(#1\right]}

\let\oldnl\nl
\newcommand{\nolinenumber}{\renewcommand{\nl}{\let\nl\oldnl}}

%% file: sections/0b_acronyms.tex
\newacronym{grstaps}{GRSTAPS}{Graphically Recursive Simultaneous Task Allocation, Planning, and Scheduling}
\newacronym{stapstc}{STAP-STC}{Simultaneous Task Allocation and Planning with Spatiotemporal Constraints}
\newacronym{stpa}{STPA}{Sequential Task Planning Anytime}
\newacronym{staa}{STAA}{Sequential Task Allocation Anytime}

\newacronym{stn}{STN}{Simple Temporal Network}
\newacronym{stp}{STP}{Simple Temporal Problem}
\newacronym{pstn}{PSTN}{Probabilistic Simple Temporal Network}
\newacronym{pstp}{PSTP}{Probabilistic Simple Temporal Problem}
\newacronym{rcpsp}{RCPSP}{Resource Constrained Project Scheduling Problem}
\newacronym{hcs}{HCS}{Heterogeneous Coalition Scheduling}
\newacronym{hcstu}{HCSTU}{Heterogeneous Coalition Scheduling with Temporal Uncertainty}
\newacronym{hcsd}{HCSD}{Heterogeneous Coalition Scheduling with Deadlines}
\newacronym{cshssrg}{CS-HSSRG}{Coalition Scheduling with Heuristic Sample Selection and Risk Guarantee}

\newacronym{mrss}{MRSs}{Multi-Robot Systems}
\newacronym{mrs}{MRS}{Multi-Robot System}
\newacronym{hmrs}{HMRS}{Heterogeneous Multi-Robot Systems}

\newacronym{mrta}{MRTA}{Multi-Robot Task Allocation}
\newacronym{gbfs}{G-BFS}{Greedy Best-First Search}
\newacronym{itags}{ITAGS}{Incremental Task Allocation Graph Search}
\newacronym{ctas}{CTAS}{Capability-based Robust Task Assignment and Scheduling}
\newacronym{apr}{APR}{Allocation Percentage Remaining}
\newacronym{nsq}{NSQ}{Normalized Schedule Quality}
\newacronym{tetaq}{TETAQ}{Time-Extended Task Allocation Quality}
\newacronym{traits}{TRAITS}{\textbf{TR}ait Throughput and \textbf{A}llocation \textbf{I}nformed \textbf{T}eam \textbf{S}cheduling}

\newacronym{madla}{MADLA}{Multi-Agent Distributed and Local Asynchronous}
\newacronym{psm}{PSM}{Planning State Machine}
\newacronym{fmap}{FMAP}{Forward Multi-Agent Planning}
\newacronym{mafs}{MAFS}{Multi-agent Forward Search}
\newacronym{pomp}{POMP}{Partial-Order Multi-agent Planning}
\newacronym{spa}{SPA}{Single-agent Planning Approach}
\newacronym{mapja}{MAPJA}{Multi-Agent Planning with Joint Actions}

\newacronym{mmmp}{MMMP}{Multi-Modal Motion Planning}
\newacronym{tsp}{TSP}{Travelling Salesman Problem}
\newacronym{lpp}{LPP}{Longest Path Problem}

\newacronym{ompl}{OMPL}{Open Motion Planning Library}
\newacronym{prm}{PRM}{Probabilistic Roadmap}
\newacronym{rrt}{rrt}{Rapidly-Exploring Random Trees}
\newacronym{ctp}{CTP}{Canadian Traveler Problem}
\newacronym{cbs}{CBS}{Conflict-Based Search}
\newacronym{mrmp}{MRMP}{Multi-Robot Motion Planning}

\newacronym{mvr}{MVR}{Multi-Vehicle Routing}
\newacronym{mamp}{MAMP}{Multi-Agent Motion Planning}
\newacronym{mvrptw}{MVRP-TW}{Multi-Vehicle Routing Problem with Time Windows}
\newacronym{cfstp}{CFSTP}{Coalition Formation with Spatial and Temporal Constraints Problem}

\newacronym{cbfs}{CBFs}{Control Barrier Functions}

\newacronym{sprt}{SPRT}{Sequential Probability Ratio Test}
\newacronym{ehc}{EHC}{Enhanced Hill Climbing}
\newacronym{ltl}{LTL}{Linear Temporal Logic}
\newacronym{stl}{STL}{Signal Temporal Logic}
\newacronym{mitl}{MITL}{Metric Interval Temporal Logic}
\newacronym{catl}{CaTL}{Capability Temporal Logic}
\newacronym{map}{MAP}{Multi-Agent Planning}
\newacronym{tamp}{TAMP}{Task and Motion Planning}
\newacronym{smt}{SMT}{Satisfiability Modulo Theorem}
\newacronym{nlp}{NLP}{Nonlinear Linear Program}
\newacronym{milp}{MILP}{Mixed-Integer Linear Programming}
\newacronym{mip}{MIP}{Mixed-Integer Programming}
\newacronym{miqp}{MIQP}{Mixed-Integer Quadratic Programming}
\newacronym{stap}{STAP}{Simultaneous Task Allocation and Planning}
\newacronym{mdp}{MDP}{Markov Decision Process}
\newacronym{saa}{SAA}{Sample Average Approximation}
\newacronym{lp}{LP}{Linear Programming}
\newacronym{cvar}{C-Var}{Conditional Value-At-Risk}

%% file: sections/0c_math_symbols.tex
\newcommand{\R}{\ensuremath{\mathbb{R}}} 

\newcommand{\species}{\ensuremath{\psi}}

\newcommand{\numRobots}{\ensuremath{N}}
\newcommand{\robot}{\ensuremath{r}}

\newcommand{\robotIndex}{\ensuremath{n}}
\newcommand{\robotIndexSet}{\ensuremath{\mathcal{\numRobots}}}

\newcommand{\numTasks}{\ensuremath{K}}
\newcommand{\task}{\ensuremath{\tau}}

\newcommand{\taskI}[1][\taskIndex]{\ensuremath{\task_{#1}}}

\newcommand{\taskIndex}{\ensuremath{k}}
\newcommand{\taskIndexSet}{\ensuremath{\mathcal{\numTasks}}}

\newcommand{\taskStartTime}[1][\taskIndex]{\ensuremath{t_{s}^{#1}}}
\newcommand{\taskEndTime}[1][\taskIndex]{\ensuremath{t_{f}^{#1}}}

\newcommand{\Trait}{\ensuremath{Q}}
\newcommand{\trait}{\ensuremath{q}}

\newcommand{\traitRate}{\ensuremath{\dot{\trait}}}

\newcommand{\numTraits}{\ensuremath{U}}
\newcommand{\traitIndex}{\ensuremath{u}}
\newcommand{\traitIndexSet}{\ensuremath{\mathcal{\numTraits}}}

\newcommand{\stateVar}{\ensuremath{s}}

\newcommand{\initialStateK}[1][\taskIndex]{\ensuremath{\stateVar_{i}^{#1}}}

\newcommand{\terminalStateK}[1][\taskIndex]{\ensuremath{\stateVar_{t}^{#1}}}

\newcommand{\initialConfigurations}{\ensuremath{I_c}}
\newcommand{\initialConfigurationsN}[1][\robotIndex]{\ensuremath{\initialConfigurations^{(#1)}}}
\newcommand{\motionPlanI}[1][\taskI]{\ensuremath{\pi}}
\newcommand{\motionPlans}{\ensuremath{\Pi}}

\newcommand{\makespan}{\ensuremath{C}}
\newcommand{\makespanLB}{\ensuremath{\makespan_{lb}}}
\newcommand{\makespanUB}{\ensuremath{\makespan_{ub}}}

\newcommand{\schedule}{\ensuremath{\sigma}}

\newcommand{\precedenceConstraints}{\ensuremath{\mathcal{P}}}

\newcommand{\mutexIndicator}{\ensuremath{\delta}}
\NewDocumentCommand{\mutexIndicatorIJ}{ O{i} O{j} }{\ensuremath{\mutexIndicator_{#1#2}}}
\newcommand{\mutexConstraints}{\ensuremath{\mathcal{M}}}

\newcommand{\absoluteDeadlineConstraints}{\ensuremath{T_{abs}}}
\newcommand{\relativeDeadlineConstraints}{\ensuremath{T_{rel}}}

\newcommand{\allocation}{\ensuremath{\mathbf{A}}}
\newcommand{\allocationKN}{\ensuremath{A_{\taskIndex \robotIndex}}}

\newcommand{\desiredTraits}[1][*]{\ensuremath{{Y}^{#1}}} 
\newcommand{\desiredTraitsMatrix}[1][*]{\ensuremath{\mathbf{Y}^{#1}}}

\NewDocumentCommand{\transitionDurationIJ}{ O{i} O{j} }{\ensuremath{\transitionDuration_{#1#2}}}
\NewDocumentCommand{\transitionDurationJI}{ O{j} O{i} }{\ensuremath{\transitionDuration_{#1#2}}}
\NewDocumentCommand{\transitionDurationIJMax}{ O{i} O{j} }{\ensuremath{\transitionDuration^{#1#2}_{\max}}}

\newcommand{\init}{\ensuremath{{0}}}
\newcommand{\intra}{\ensuremath{{\mapsto}}}
\newcommand{\inter}{\ensuremath{{\leftrightharpoons}}}

\newcommand{\inexhaustibleTrait}{IET}
\newcommand{\exhaustibleTrait}{ET}
\newcommand{\nonProvisioningTrait}{NPT}
\newcommand{\instantaneousProvisioningTrait}{IPT}
\newcommand{\gradualProvisioningTrait}{GPT}
\newcommand{\nonCumulativeTrait}{NCT}
\newcommand{\cumulativeTrait}{CT}
\newcommand{\inexhaustibleTraits}{IETs}
\newcommand{\exhaustibleTraits}{ETs}
\newcommand{\nonProvisioningTraits}{NPTs}
\newcommand{\instantaneousProvisioningTraits}{IPTs}

\newcommand{\nonCumulativeTraits}{NCTs}
\newcommand{\cumulativeTraits}{CTs}

\newcommand{\allocatedRobots}[1][\taskIndex]{\ensuremath{A_{\robot}\parens{{#1}}}}

\newcommand{\allocatedTasks}[1][\robotIndex]{\ensuremath{A_{\task}\parens{{#1}}}}

\NewDocumentCommand{\robotTraitInitNU}{O{(\robotIndex)} O{,\traitIndex}}{
  \ensuremath{\Trait_{\init #2}^{#1}}}

\NewDocumentCommand{\desiredTraitsKU}{O{\taskIndex} O{\traitIndex}}{
  \ensuremath{\desiredTraits_{#1#2}}}

\newcommand{\robottrait}{\ensuremath{\trait}}
\newcommand{\robotTrait}{\ensuremath{\Trait}}

\NewDocumentCommand{\robottraitKNU}{O{\taskIndex} O{(\robotIndex)} O{\traitIndex}}{
  \ensuremath{\robottrait_{#1#3}^{#2}}}
\NewDocumentCommand{\robotTraitKNU}{O{\taskIndex} O{(\robotIndex)} O{\traitIndex}}{
  \ensuremath{\robotTrait_{#1#3}^{#2}}}

\NewDocumentCommand{\robotTraitRateInitNU}{O{(\robotIndex)} O{,\traitIndex}}{
  \ensuremath{\dot{\Trait}_{{0} #2}^{#1}}}
\NewDocumentCommand{\robotMaxTraitRateNU}{O{(\robotIndex)} O{,\traitIndex}}{
  \ensuremath{\dot{\Trait}_{{\max} #2}^{#1}}}

\newcommand{\robottraitRate}{\ensuremath{\dot{\robottrait}}} 
\newcommand{\robotTraitRate}{\ensuremath{\dot{\robotTrait}}}

\NewDocumentCommand{\robottraitRateKNU}{O{\taskIndex} O{(\robotIndex)} O{\traitIndex}}{\ensuremath{\robottraitRate_{#1#3}^{#2}}}
\NewDocumentCommand{\robotTraitRateKNU}{O{\taskIndex} O{(\robotIndex)} O{\traitIndex}}{\ensuremath{\robotTraitRate_{#1#3}^{#2}}}
  
\newcommand{\traitMismatch}{\ensuremath{E}}
\newcommand{\traitRateMismatch}{\ensuremath{\dot{\traitMismatch}}}
\NewDocumentCommand{\traitMismatchKU}{O{\taskIndex} O{\traitIndex}}{\ensuremath{\traitMismatch_{#1#2}}}
\NewDocumentCommand{\traitRateMismatchKU}{O{\taskIndex} O{\traitIndex}}{\ensuremath{\traitRateMismatch_{#1#2}}}

\NewDocumentCommand{\desiredTraitRatesKU}{O{\taskIndex} O{\traitIndex}}{\ensuremath{\dot{Y}^{*}_{#1#2}}}
\newcommand{\desiredTraitRatesMatrix}{\ensuremath{\dot{\mathbf{Y}}^{*}}}

\NewDocumentCommand{\coalitionTraitKU}{O{\taskIndex} O{\traitIndex}} {\ensuremath{y_{#1 #2}}}

\NewDocumentCommand{\coalitionTraitRateKU}{O{\taskIndex} O{\traitIndex}}
{\ensuremath{\dot{y}_{#1 #2}}}

\newcommand{\transitionLength}{\ensuremath{l}}

\newcommand{\initialTransitionLengthN}[1][\robotIndex]{\ensuremath{\transitionLength \left(\initialConfigurationsN, \initialStateK[1], \species(\robotIndex) \right)}}

\newcommand{\interTransitionLengthN}[1][\robotIndex]{\ensuremath{\transitionLength}_{\inter}^{(#1)}}

\newcommand{\interTransitionLengthTrueN}[1][\robotIndex]{\ensuremath{\transitionLength}_{\inter}^{(#1)}}

\newcommand{\interTransitionLengthEstimateN}[1][\robotIndex]{\ensuremath{\hat{\transitionLength}}_{\inter}^{(#1)}}

\newcommand{\taskDuration}{\ensuremath{d}}
\newcommand{\taskDurationI}[1][\taskIndex]{\ensuremath{d_#1}} 
\newcommand{\taskDurationK}[1][\taskIndex]{\ensuremath{\taskDuration_#1}}
\newcommand{\taskDurationKS}[1][\taskIndex]{\ensuremath{\taskDuration_{#1}^{s}}}
\newcommand{\taskDurationKQ}[1][\taskIndex]{\ensuremath{\taskDuration_{#1}^{\trait}}}

\newcommand{\taskDurationKUQ}[1][\taskIndex]{\ensuremath{\taskDuration_{#1 \traitIndex}^{\trait}}}

\newcommand{\taskDurationKNUQ}[1][\taskIndex]{\ensuremath{\taskDuration_{#1u}^{q(n)}}}

\newcommand{\transitionDuration}{\ensuremath{\phi}}

\newcommand{\initialTransitionDuration}{\ensuremath{\transitionDuration_\init}}
\newcommand{\initialTransitionDurationK}[1][\taskIndex]{\ensuremath{\initialTransitionDuration^{#1}}}

\newcommand{\intraTransitionDuration}[1][\taskIndex]{\ensuremath{\transitionDuration^{#1}_{\intra}}}
\newcommand{\interTransitionDuration}{\ensuremath{\transitionDuration}_{\inter}}
\newcommand{\interTransitionDurationN}[1][\robotIndex]{\ensuremath{\interTransitionDuration^{(#1)}}}

\newcommand{\batteryCurrent}{\ensuremath{I}}

\NewDocumentCommand{\batteryIdleCurrentN}{O{(\robotIndex)}}{\ensuremath{\batteryCurrent_{idle}^{#1}}}
\NewDocumentCommand{\batteryCurrentKNU}{O{\taskIndex} O{(\robotIndex)} O{\traitIndex}}{\ensuremath{\batteryCurrent_{#1#3}^{#2}}}
\newcommand{\batteryVoltageN}[1][\robotIndex]{\ensuremath{V^{(#1)}}}
\newcommand{\maxBatteryCurrentN}{\ensuremath{\batteryCurrent_{\max}^{(\robotIndex)}}}
\NewDocumentCommand{\traitCurrentCoeffNU}{O{(\robotIndex)} O{\traitIndex}}{\ensuremath{c^{#1}_{\trait, #2}}}
\NewDocumentCommand{\traitRateCurrentCoeffNU}{O{(\robotIndex)} O{\traitIndex}}{\ensuremath{c^{#1}_{\traitRate, #2}}}
\NewDocumentCommand{\velocityCurrentCoeffN}{O{(\robotIndex)}}{\ensuremath{c^{#1}_{\velocity}}}
\newcommand{\PeukertExponent}[1][(\robotIndex)]{\ensuremath{p^{#1}}}

\newcommand{\Battery}{\ensuremath{B}}
\newcommand{\battery}{\ensuremath{b}}
\newcommand{\batteryInit}{\ensuremath{\Battery_\init}}
\newcommand{\batteryInitN}[1][\robotIndex]{\ensuremath{\batteryInit^{(#1)}}}
\newcommand{\batteryN}[1][\robotIndex]{\ensuremath{\battery^{(#1)}}}
\NewDocumentCommand{\batteryKNU}{O{\taskIndex} O{(\robotIndex)} O{\traitIndex}}{\ensuremath{\battery_{#1#3}^{#2}}}
\newcommand{\batteryInterN}[1][\robotIndex]{\ensuremath{\battery^{(#1)}_{\inter}}}

\newcommand{\velocity}{\ensuremath{\nu}}
\newcommand{\velocityKN}[1][\robotIndex]{\ensuremath{\nu_{\taskIndex}^{(#1)}}}
\newcommand{\velocityInter}[1][\robotIndex]{\ensuremath{\nu_{\inter}}}
\newcommand{\velocityInterN}[1][\robotIndex]{\ensuremath{\nu^{(#1)}_{\inter}}}
\newcommand{\maxVelocityN}[1][\robotIndex]{\ensuremath{\nu_{\max}^{(#1)}}}

\newcommand{\coalitionIntraVelocityK}{\ensuremath{\nu_{\intra}^{\taskIndex}}}

\newcommand{\traitProvisions}{\ensuremath{\robottrait}}
\newcommand{\traitProvisionRates}{\ensuremath{\dot{\robottrait}}}

\newcommand{\traitsSolution}{\ensuremath{\angled{\allocation, \traitProvisions, \traitProvisionRates, \schedule, \motionPlans}}}


%% file: sections/0d_title.tex
\title{Modeling and Optimizing the Provisioning of Exhaustible Capabilities for Simultaneous Task Allocation and Scheduling}

\author{Jinwoo Park}
\orcid{0000-0002-0318-2562}
\affiliation{
  \institution{Georgia Institute of Technology}
  \city{Atlanta}
  \state{GA}
  \country{USA}}
\email{jinwoop@gatech.edu}

\author{Harish Ravichandar}
\orcid{0000-0002-6635-2637}
\affiliation{
  \institution{Georgia Institute of Technology}
  \city{Atlanta}
  \state{GA}
  \country{USA}}
\email{harish.ravichandar@cc.gatech.edu}

\author{Seth Hutchinson}
\orcid{0000-0002-3949-6061}
\affiliation{
  \institution{Northeastern University}
  \city{Boston}
  \state{MA}
  \country{USA}}
\email{s.hutchinson@northeastern.edu}

%% file: sections/1_abstract.tex
Deploying heterogeneous robot teams to accomplish multiple tasks over extended time horizons presents
significant computational challenges for task allocation and planning.
In this paper, we present a comprehensive, time-extended, offline heterogeneous multi-robot task allocation framework,
\acrshort*{traits}, which we believe to be the first that can cope with the provisioning of exhaustible traits under battery and temporal constraints.
Specifically, we introduce a nonlinear programming-based trait distribution module that can optimize the trait-provisioning rate of coalitions to yield feasible and time-efficient solutions.
\acrshort*{traits} provides a more accurate feasibility assessment and estimation of task execution times and makespan by leveraging trait-provisioning rates while optimizing battery consumption—an advantage that state-of-the-art frameworks lack.
We evaluate \acrshort*{traits} against two state-of-the-art frameworks, with results demonstrating its advantage in satisfying complex trait and battery requirements while remaining computationally tractable.

%% file: sections/2_introduction.tex
\section{Introduction}
Given their ability to collectively execute tasks, \acrfull*{mrss} enhance productivity and robustness~\cite{dias2004robust}, leading to their increasing deployment in logistics, agriculture, and manufacturing. 
Heterogeneous \acrshort*{mrss}, composed of robots with different \textit{traits} (capabilities), offer greater flexibility and efficiency in multi-agent coordination, particularly when tasks involve complex requirements that cannot be satisfied by homogeneous agents~\cite{shehory1998methods}.
Such systems must simultaneously reason about the intertwined problems of task allocation (\textit{who}), scheduling (\textit{when}), and motion planning (\textit{how}) while satisfying multidimensional task requirements~\cite{neville2023ditags}.
However, identifying even a feasible~plan for this problem is NP-hard due to its combinatorial nature~\cite{messing2022grstaps}. 

While prior works have proven effective in handling challenges associated with coordinating heterogeneous \acrshort*{mrss}, they ignore crucial practical considerations that exacerbate computational challenges. 
See \figref{warehouse} for an illustrative example.
First, robot traits may be either \textit{exhaustible} (e.g., deployable resources) or \textit{inexhaustible} (e.g., sensing radius). Ignoring the exhaustible nature of traits can lead to overly optimistic solutions that fail to achieve adequate throughput.
Second, tasks might require that exhaustible traits be \textit{provisioned} at a particular rate (e.g., chemical dispensing). Ignoring provisioning can lead to suboptimal task performance or failures.
Third, it is necessary to account for \textit{energy} consumption (e.g., battery drain). 
Plans that ignore energy considerations risk being overly optimistic and violating important constraints such as battery capacity or fuel supply.
Together, these factors considerably expand the search space, increase the computational burden, and complicate modeling and optimization.

\begin{figure}[t]
    \centering
    \includegraphics[width=0.99\linewidth, trim={2mm, 2mm, 2mm, 2mm}, clip]{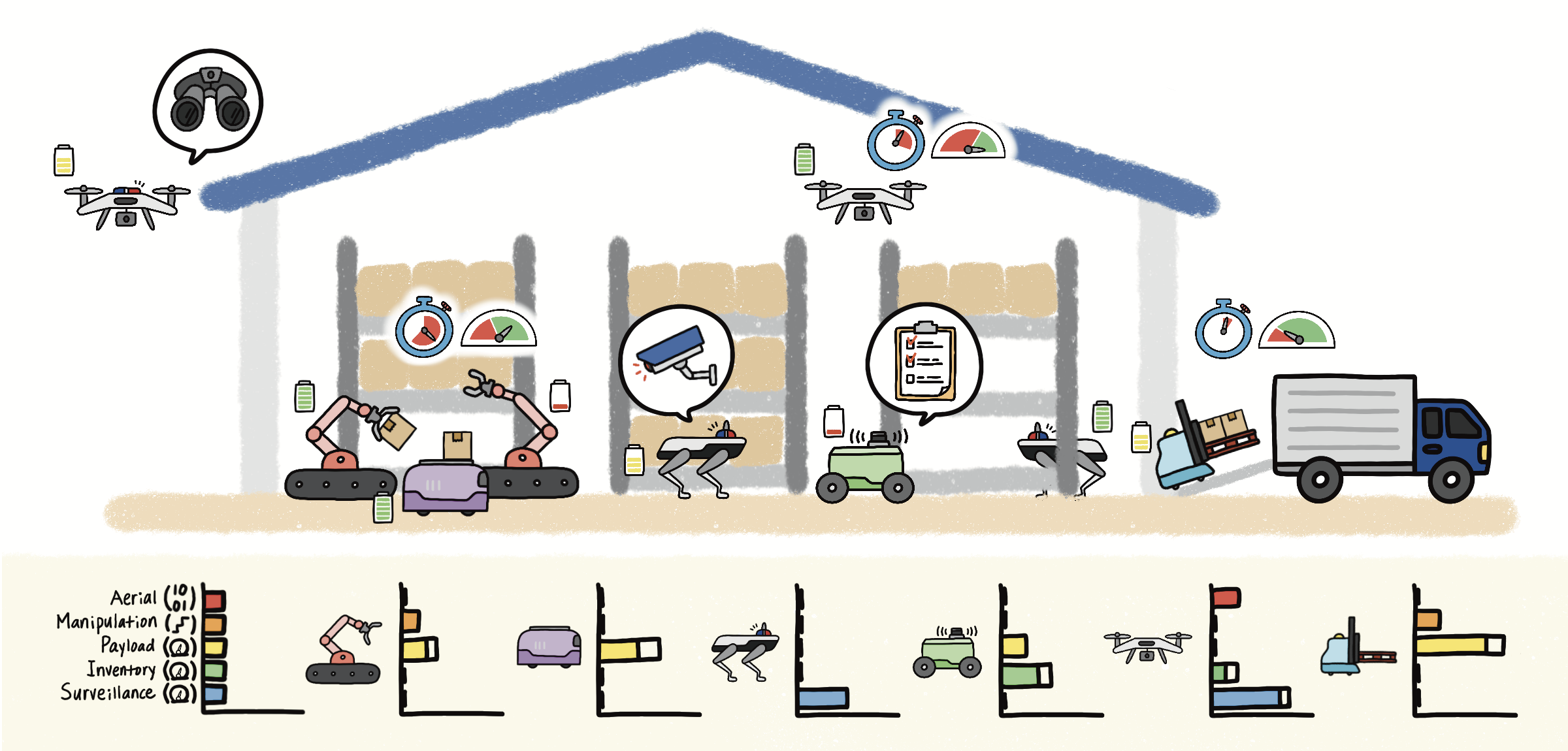}
    \caption{ 
        A motivating scenario: An autonomous warehouse with heterogeneous robots (top), characterized by traits that have provisioning and exhaustibility attributes (bottom), collaborating on complex tasks subject to temporal and battery constraints.
    }
    \Description{A motivating scenario figure.}
    \figlabel{warehouse}
\end{figure}

In this work, we introduce 
\textit{\acrlong*{traits}}~(\acrshort*{traits}), a new modeling and algorithmic framework that extends the capabilities of existing heterogeneous \acrfull*{mrta} approaches to include the dynamics of exhaustible traits and their allocation within teams operating under battery constraints.
Introducing trait-provisioning rates and battery optimization facilitates more accurate estimation of task durations and sustained robot operations.
To this end, \acrshort*{traits} simultaneously solves the trait-provisioning problem to generate a plan that satisfies multiple constraints.

In summary, we contribute
\begin{enumerate}[label=(\roman*), itemsep=0pt, topsep=0pt]
    \item a novel modeling framework that formalizes the provisioning problem for exhaustible traits, while maintaining compatibility with existing approaches to trait-based task allocation,
    \item a comprehensive time-extended task allocation algorithm capable of handling traits with attributes of exhaustibility, provisioning, and varying forms of aggregation,
    \item an offline \acrshort*{mrta} framework that models battery usage by linking trait provision to current draw, incorporating Peukert's law to optimize robot utilization.
\end{enumerate}

We evaluated the performance of \acrshort*{traits} across various scenarios in simulated warehouse environments with different trait requirements.
We compared \acrshort*{traits} against two state-of-the-art heterogeneous \acrshort*{mrta} frameworks that can only handle a subset of traits that \acrshort*{traits} can manage and do not incorporate battery considerations.
The results show that \acrshort*{traits} can effectively handle a broader range of traits and generate battery-optimized, feasible plans with computational tractability.
The source code is publicly available at \url{https://github.com/GT-STAR-Lab/TRAITS}.

%% file: sections/3_related_work.tex
\section{Related Work}

The robot coordination problem is typically categorized into instantaneous allocation (IA) and time-extended allocation (TA)~\cite{gerkey2004formal}. Numerous IA frameworks for heterogeneous task allocation have demonstrated high computational efficiency in coordinating robots under various compounding factors, such as energy~\cite{mayya2021resilient, notomista2022resilient}, distributed optimization~\cite{shorinwa2023distributed}, differing capabilities and resources~\cite{shiroma2009comutar, xu2022resource}, making them well-suited for online computation and practical deployment.
However, these approaches lack the ability to reason about task execution order and are incapable of producing viable solutions when the number of robots is fewer than the number of tasks.
In contrast, many TA frameworks simultaneously address the combinatorial coordination problem while additionally accounting for uncertainties~\cite{fu2023robust, neville2023ditags, messing2023sampling}, tighter temporal constraints~\cite{park2023ritags, zhang2025energy} alongside the increased complexity of scheduling in a higher-dimensional search space.  
Nevertheless, these frameworks typically assume time-invariant resource availability.

\acrshort*{mrta} architectures must cope with numerous criteria, such as competence, energy, resource, spatial, and temporal constraints.
The market-based~\cite{koenig2006power, choi2009consensus, choi2010decentralized} and optimization-based~\cite{prorok2017impact, gombolay2018tercio, leahy2022scratches, aswale2023heterogeneous} approaches have effectively addressed task allocation problems.
Furthermore, probabilistic~\cite{choudhury2022dynamic}, game-theoretical~\cite{chen2023game}, and learning~\cite{banfi2022hierarchical, srikanthan2022resource} methods were proposed to tackle interdependent subproblems simultaneously.
Despite their novelties and capability of finding feasible solutions, both resource and energy management were outside the scope of their research.

Resource management under energy constraints for \acrshort*{mrs} has been the focus of various research efforts. However, the methods that account for energy either involve instantaneous allocations without scheduling~\cite{dressler2005energy, notomista2022resilient}, or abstract energy consumption in terms of time~\cite{calvo2024optimal}.
Moreover, the resource constraint problem addressed in~\cite{beynier2007decentralized, schillinger2017multi, chen2011resource} has either been demonstrated only with a small set of robots or does not account for energy constraints.
A decentralized, auction-based approach~\cite{lee2018resource} effectively managed resources and battery consumption, though it was limited to homogeneous robots. 
In contrast, the algorithm for heterogeneous coalitions~\cite{wu2017gini} does not require all tasks to be completed.

Recent works~\cite{ravichandar2020strata, neville2021interleaved, fu2023robust, neville2023ditags, park2023ritags, ito2025energy, liu2026ijrr} leverage the notion of \textit{traits} to address coordination problems in heterogeneous multi-robot systems. 
However, these approaches have modeled only traits that are inexhaustible and instantaneously provisioned.
In particular, studies \cite{neville2021interleaved, neville2023ditags, park2023ritags, fu2023robust, ito2025energy, liu2026ijrr} consider only inexhaustible traits that are either non-provisioned or provisioned instantaneously.
Although \cite{fu2023robust} and \cite{gosrich2023multi} formulate task allocation as a network-flow problem, task duration is still treated as constant.
The concept of \textit{resource-spending velocity}, or \textit{resource burn rate} was introduced in~\cite{kaminka2010adaptive} from a game-theoretic perspective, employing reinforcement learning to address decentralized, time-extended multi-robot foraging problem.
However, all robots were assumed to be homogeneous and relied solely on fuel as their resource.
The concept of a dynamic trait model, where trait values can change, was introduced in~\cite{neville2020approximated}; however, traits are ultimately transformed into constant values for computational purposes.
The approach in~\cite{ito2025energy} incorporates robot capabilities, collision avoidance, and energy constraints; however, it does not account for scheduling and trait exhaustibility.
While this trait-based \acrshort*{mrta} research has introduced novel approaches to address challenging spatio-temporal constraints, these studies still assume traits to be constant, without incorporating time-dependent rates.
A shortcoming of assuming constant traits is that such solutions cannot capture the reduction in task duration that occurs when additional robots are added to a coalition.

%% file: sections/4_problem_description.tex
\section{TRAITS: Modeling Framework}
Our problem of interest is a variant of XD\,[ST-MR-TA]:SP and TW problems, defined by the established MRTA taxonomies~\cite{gerkey2004formal, korsah2013comprehensive, nunes2017taxonomy}.
We consider scenarios with $\numTasks$ tasks and $\numRobots$ robots.
Each robot's capabilities are defined by a distinct trait composition, drawn from a total of 
$\numTraits$ traits, where the set and magnitude of traits vary across robots.
Let us define the index sets for \textit{robots}, \textit{tasks}, and \textit{traits} as 
$\robotIndexSet, \taskIndexSet$, and $\traitIndexSet$,
respectively. 

\textit{Problem Statement}: 
Given a set of robot traits, desired provisioning rates, battery capacities, and operational constraints, the objective is to determine a feasible plan that specifies how traits should be provisioned across tasks in a way that minimizes overall task execution time and satisfies task specifications while honoring battery constraints.

To address the above problem, we first introduce our comprehensive new modeling framework in this section. We begin with formal definitions of fundamental quantities to capture the notion of trait provisioning (Sec. \ref{sec:provisioning_model}), then discuss our new model and expanded taxonomy of traits (Sec. \ref{sec:a_new_trait_taxonomy}), followed by a discussion of how our model influences trait aggregation (Sec. \ref{sec:trait_aggregation}) and scheduling (Sec. \ref{sec:task_and_transition_duration}), and conclude the section with a model of how battery consumption can impact task allocation (Sec. \ref{sec:battery_consumption}).

\subsection{Modeling Trait Provisioning}
\label{sec:provisioning_model}
While past research has treated traits as irreducible quantities that can be instantly deployed~\cite{messing2022grstaps, prorok2017impact, ravichandar2020strata}, we reformulate them as \textit{dynamic} quantities that need to be regulated over time and optimized based on task-specific constraints.

We denote the initial traits (e.g., resource, sensing radius) and the maximum trait-provisioning rates of the $\robotIndex$-th robot by $\robotTraitInitNU[(\robotIndex)][] \in \R_{\geq 0}^{1 \times \numTraits}$ and $\robotMaxTraitRateNU[(\robotIndex)][] \in \R_{\geq 0}^{1 \times \numTraits}$, respectively.
The scalar entries $\robotTraitInitNU$~and $\robotMaxTraitRateNU$~represent the initial amount and the maximum provisioning rate of the $\traitIndex$-th trait, respectively.

Let the team be collectively responsible for completing $\numTasks$ tasks. 
The desired trait matrix $\desiredTraitsMatrix \in \R_{\geq 0}^{\numTasks \times \numTraits}$ and desired trait-provisioning rate matrix $\desiredTraitRatesMatrix \in \R_{\geq 0}^{\numTasks \times \numTraits}$ specify the minimum trait quantities and provisioning rates required to successfully execute all tasks, respectively.
We use $\desiredTraitsKU$ and $\desiredTraitRatesKU$ to denote the required $\traitIndex$-th trait and its provisioning rate for the $\taskIndex$-th task, respectively.

The core of our problem is to determine \emph{how each trait should be provisioned} by each robot, subject to the definitions above.
Let $\robottraitKNU$
represent a decision variable denoting how much of $\robotIndex$-th robot's $\traitIndex$-th trait should be provisioned for the $\taskIndex$-th task, yielding a decision vector for each robot:
\begin{equation*}
    \robottraitKNU[\taskIndex][(\robotIndex)][] = 
    \begin{bmatrix}
        \robottraitKNU[\taskIndex][(\robotIndex)][1], 
        \robottraitKNU[\taskIndex][(\robotIndex)][2], 
        \ldots, 
        \robottraitKNU[\taskIndex][(\robotIndex)][\numTraits]
    \end{bmatrix}
    \in \R^{1 \times \numTraits}_{\geq 0}
    \quad \forall \robotIndex \in \robotIndexSet
    \; \forall \taskIndex \in \taskIndexSet
    .
\end{equation*}
$\robottraitKNU \in \R_{\geq 0}$ for continuous or discrete traits, and $\robottraitKNU \in \{0, 1\}$ for binary traits.
If the robot lacks the trait, then $\robottraitKNU = 0$.

Similarly, we define a trait-provisioning rate vector $\robottraitRateKNU[\taskIndex][(\robotIndex)][] \in \R_{\geq 0}^{1 \times \numTraits}$ as a decision variable which represents the rate at which the $\robotIndex$-th robot must provide its traits for the $\taskIndex$-th task.
The provisioning rate $\robottraitRateKNU$ is positive only when the trait can be provisioned gradually; otherwise, it is zero.

\subsection{A New Trait Taxonomy}
\label{sec:a_new_trait_taxonomy}
Multi-robot tasks (MR) require multiple traits, where some traits support cooperative contribution from the coalition, while others demand a minimum level of proficiency from each individual robot. 
Additionally, the total contribution from any robot across tasks may be limited by its initial trait capacity.
Despite their practical significance, no existing taxonomy or model captures these considerations~\cite{ravichandar2020strata, neville2021interleaved, fu2023robust}.

\begin{figure}[tbh]
    \centering
    \includegraphics[width=0.8\linewidth]{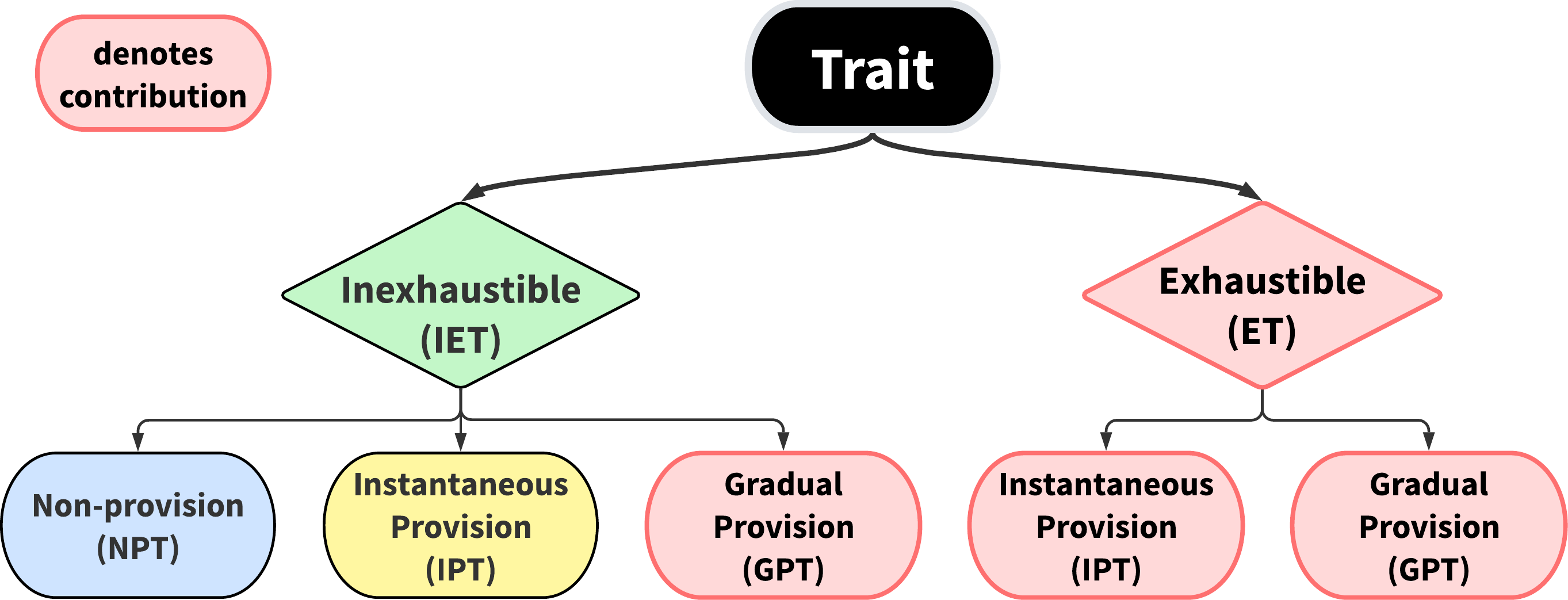}
    \caption{
        The proposed model captures various trait aspects and the framework handles all categories in the diagram.
    }
    \figlabel{trait_taxonomy}
    \Description{Figure demonstrates different branches of trait types.}
\end{figure}

Based on the definitions in Sec. \ref{sec:provisioning_model}, we propose a new taxonomy for traits (see \figref{trait_taxonomy}):

\subsubsection{Exhaustible Traits}
The exhaustibility attribute of a trait indicates whether the total provision of a robot’s traits across multiple tasks is limited by its initial trait levels. 
An \textbf{inexhaustible trait (\inexhaustibleTrait)} is not depleted through use and typically represents inherent properties or infinitely replicable entities. 
In contrast, an \textbf{exhaustible trait (\exhaustibleTrait)} is consumed with each task and generally corresponds to quantitatively measurable, finite resources carried by the robot.
Examples of \inexhaustibleTraits~include maximum payload capacity, monitoring capability, and sensor range, whereas examples of \exhaustibleTraits~include the volume of chemicals, battery level, and other consumable resources.
Thus, the following constraint must hold: 
\begin{equation}
    \robotTraitInitNU \geq
    \begin{cases}
        \max\limits_{\taskIndex \in \allocatedTasks} \robottraitKNU & \text{if} \; \inexhaustibleTrait, \\
        \sum\limits_{\taskIndex \in \allocatedTasks} \robottraitKNU & \text{if} \; \exhaustibleTrait,
    \end{cases}
    \equlabel{exhaustible_trait_definition}    
\end{equation}
\noindent
where \allocatedTasks~is the set of task indices allocated to the $\robotIndex$-th robot.

\subsubsection{Provisioning Traits}
A provisioning attribute of a trait indicates whether it can be provisioned and, if so, defines the feasible provisioning rate. 
As such, we categorize traits into three groups:
\begin{enumerate}[label=(\roman*), itemsep=0pt, topsep=0pt]
    \item \textbf{Non-provisioning traits (\nonProvisioningTrait s)}: These traits represent constant or qualitative properties (e.g., robot type, regulation compliance) that cannot be provisioned.
    \item \textbf{Instantaneous-provisioning traits (\instantaneousProvisioningTrait s)}: These traits can be transferred instantaneously 
    (i.e., $\robottraitRateKNU = \infty$),
    such as occupying a seat, engaging an emergency stop, and transferring a token. They are generally discrete, single, or one-time-use in nature.
    \item \textbf{Gradual-provisioning traits (\gradualProvisioningTrait s)}: These traits require time to be transferred, i.e., $\robottraitRateKNU \in \left(0, \robotMaxTraitRateNU \right] $, and typically represent continuous processes such as gas refueling or water provisioning for fire suppression.
\end{enumerate}
To make the problem tractable, we define
the gradual provisioning rate $\robottraitRateKNU$ as the constant rate at which the $\robotIndex$-th robot provides the $\traitIndex$-th trait to complete the $\taskIndex$-th task during its execution. 
The desired trait-provisioning rate is set to zero $\parens{\desiredTraitRatesKU = 0}$ for all \nonProvisioningTraits~and \instantaneousProvisioningTraits.

Whether a trait is continuous or discrete depends on the level of granularity in its definition---such as whether it is considered a distinct unit, a flow, or a bulk entity.  
Moreover, certain traits, such as the LiDAR scanning angle, exhibit an inexhaustible nature yet may impose temporal requirements, such as the scanning frequency. 
These can be modeled through gradual provisioning rates, indicating that trait provision is not necessarily constrained by exhaustibility.

We define~$\taskDurationKNUQ \in \R_{\geq 0}$~to be the time required for the $\robotIndex$-th robot to apply the $\traitIndex$-th trait to the $\taskIndex$-th task 
\begin{equation}
    \taskDurationKNUQ =
    \begin{cases}
         0 \quad& \text{if} \; \nonProvisioningTrait \text{ or } \instantaneousProvisioningTrait, \\
         \robottraitKNU / \robottraitRateKNU \quad& \text{if} \; \gradualProvisioningTrait.
    \end{cases}
    \equlabel{dynamic_task_duration_definition}
\end{equation}

\subsection{Trait Aggregation}
\label{sec:trait_aggregation}
To enhance adaptability across diverse task requirements, we define aggregation functions that determine how multiple robots can make partial contributions to a collective task, using decision variables that enable task-specific adaptation.
For instance, certain tasks may necessitate the aggregation of traction force at the team level, whereas others may require ensuring a minimum traction capability at the individual level, highlighting the importance of task-specific aggregation strategies.

A {\em cumulative} attribute of a trait indicates whether a task’s required characteristic must be individually satisfied by each robot within a coalition or can be collectively achieved through the combined output of the team. We define the former as a \textbf{noncumulative trait (\nonCumulativeTrait)}, which pertains to individual robot capabilities, and the latter as a \textbf{cumulative trait (\cumulativeTrait)}, which pertains to team-level capabilities. Examples of \nonCumulativeTraits~include monitoring capability, emergency stop functionality, and the number of manipulator arms per robot. In contrast, examples of \cumulativeTraits~include traction force and storage volume, though such traits can also be modeled as \nonCumulativeTraits~depending on the task requirements.

Hence, if the $\taskIndex$-th task involves both cumulative and noncumulative traits, then for every $\traitIndex$-th trait in the $\taskIndex$-th task, we define the aggregation function
\begin{equation}
    \coalitionTraitKU =
    \begin{cases}
        \min\limits_{\robotIndex \in \allocatedRobots} \robottraitKNU & \text{if} \; \nonCumulativeTrait, \\
        \sum\limits_{\robotIndex \in \allocatedRobots} \robottraitKNU & \text{if} \; \cumulativeTrait,
        \equlabel{cummulative_trait_definition}
    \end{cases}
\end{equation}
where \allocatedRobots~is the set of indices of robots allocated to the $\taskIndex$-th task.
The time required to fulfill the $\traitIndex$-th trait of the $\taskIndex$-th task 
by robot coalition $\allocatedRobots$ is given by
$
    \taskDurationKUQ = \max\limits_{\robotIndex \in \allocatedRobots[\taskIndex]} \taskDurationKNUQ.
$
Accordingly, the simultaneous trait-provisioning rate of the coalition for the $\traitIndex$-th trait in the $\taskIndex$-th task is defined~as
\begin{equation}
    \coalitionTraitRateKU =
    \begin{cases}
         0 \quad& \text{if} \; \nonProvisioningTrait \text{ or } \instantaneousProvisioningTrait, \\
         \coalitionTraitKU / \taskDurationKUQ \quad& \text{if} \; \gradualProvisioningTrait.
    \end{cases}
    \equlabel{trait_provision_rate_definition}
\end{equation}

Non-gradual, inexhaustible provisioning traits (\nonProvisioningTrait, \instantaneousProvisioningTrait, and \inexhaustibleTrait) have been used in prior works~\cite{neville2021interleaved, messing2022grstaps, park2023ritags} to represent volume, water capacity, and maximum sensor range.
{\em
In contrast, \acrshort*{traits} is, to the best of our knowledge, the only framework that models \exhaustibleTrait~and \gradualProvisioningTrait, enabling more accurate makespan estimation and the formation of meaningful coalitions under resource and temporal constraints.
}

\subsection{Task and Transition Durations}
\label{sec:task_and_transition_duration}
The duration of the $\taskIndex$-th task is defined $\taskDurationK = \taskDurationKS + \taskDurationKQ + \intraTransitionDuration$, where $\taskDurationKS$ is the user-defined static task duration representing additional task duration unrelated to trait provisions, 
$\taskDurationKQ = \max\limits_{\traitIndex \in \traitIndexSet} \taskDurationKUQ $ is the lower bound on the time required for all trait provisions of the $\taskIndex$-th task, and $\intraTransitionDuration$ is the intra-task transition duration from the initial position $\initialStateK$ to terminal position $\terminalStateK$ of the $\taskIndex$-th task.
Furthermore, the initial transition duration for the $\taskIndex$-th task, $\initialTransitionDurationK$, represents the lower bound on the time required for all robots allocated to the $\taskIndex$-th task  
to travel from their respective initial positions $\initialConfigurations$ to $\initialStateK$.
The inter-task transition duration $\transitionDurationIJ$ is the lower bound on the time required for all robots assigned to both the $i$-th and $j$-th tasks to travel from $\terminalStateK[i]$ to $\initialStateK[j]$.

\subsection{Battery Consumption}
\label{sec:battery_consumption}
Battery consideration is essential in identifying a feasible task allocation and schedule, as robots must have enough energy to execute their assigned tasks.
A plan that considers only trait fulfillment, without accounting for energy constraints, may lead to unrealistic outcomes---such as robots being assigned an infinite number of tasks.
Rather than modeling energy as an additional $\traitIndex$-th trait, the proposed framework explicitly defines the relationship between trait values ($\robottrait, \robottraitRate$), electrical current $\batteryCurrent$ [A], and battery consumption [J]. 
Inspired by~\cite{mei2005energy}, which models power linearly with sensor frequency and motor speed, we instead model current as proportional to trait values and provisioning rates, while capturing the nonlinear relationship between current and power.
In our framework, battery current during trait delivery is modeled as a function proportional to the trait quantities and the rates at which they are provided.

The electrical current $\batteryCurrentKNU$ drawn by the $\robotIndex$-th robot to provide the $\traitIndex$-th trait for the $\taskIndex$-th task varies as a function of the trait value $\robottraitKNU$ and its provisioning rate $\robottraitRateKNU$
\begin{equation}
    \batteryCurrentKNU = 
    \traitCurrentCoeffNU \cdot \robottraitKNU
    +
    \traitRateCurrentCoeffNU \cdot \robottraitRateKNU
    \in \R,
    \equlabel{electrical_current_function}
\end{equation}
where $\traitCurrentCoeffNU \in \mathbb{R}_{\geq 0}$ and $\traitRateCurrentCoeffNU \in \mathbb{R}_{\geq 0}$ are independent coefficients representing the relationships between current draw and trait value, and between current draw and trait-provisioning rate, respectively. 
The coefficient values can be determined with domain expertise, guided by robot specification sheets (e.g., motor efficiency, torque constants, and sensor/actuator power characteristics).
Either coefficient may be set to zero, and both may be zero when the corresponding effects are negligible.
In this paper, the coefficients in the battery current model are derived from the battery’s maximum deliverable current, as specified by its C-rating, and the maximum provisioning rates or magnitudes associated with each trait. 
This ensures that current estimates remain grounded in the physical constraints of the hardware.
In our experiments, we compute these coefficients based on specifications of real-world robot platforms~\cite{clearpath_husky_specs}.

Battery current is drained even when the robot is idle $(\batteryIdleCurrentN)$ as well as when it is moving while executing a task.
The speed of this movement is referred to as the intra-task transition speed of the coalition for the $\taskIndex$-th task, $\coalitionIntraVelocityK$ that must hold
\begin{equation}
    \coalitionIntraVelocityK \leq \min\limits_{\robotIndex \in \allocatedRobots} \velocityKN \leq \min\limits_{\robotIndex \in \allocatedRobots} \maxVelocityN.
    \equlabel{intra_transition_velocity}
\end{equation}
The speed of $\robotIndex$-th robot while executing the $\taskIndex$-th task is denoted by $\velocityKN \in \sqBrackets{0, \maxVelocityN}$, and $\velocityCurrentCoeffN$ represents the relationship between battery current and robot speed.
Therefore, the average current drawn by the $\robotIndex$-th robot to accomplish the $\taskIndex$-th task can be modeled as
\begin{equation}
    \batteryCurrentKNU[\taskIndex][(\robotIndex)][]
    = \batteryIdleCurrentN + \velocityCurrentCoeffN \cdot \coalitionIntraVelocityK + \sum_{\traitIndex \in \traitIndexSet} 
    \batteryCurrentKNU,
\end{equation}
where it must hold that $\batteryCurrentKNU[\taskIndex][(\robotIndex)][] \leq \maxBatteryCurrentN$ as the current must not exceed the maximum current that the battery can supply based on the C-rating, $\maxBatteryCurrentN$.

The Peukert exponent \PeukertExponent of the $\robotIndex$-th robot is a power coefficient that characterizes how battery drain accelerates as electrical current increases.
This current increase is driven by higher trait values, greater provisioning rates, and the additional energy required for faster transitions.
Hence, the total battery consumption $\batteryN$ [J] by the $\robotIndex$-th robot must satisfy
\begin{equation}
    \batteryInitN \geq \batteryN = 
        \batteryInterN
        +        
        \sum\limits_{\taskIndex \in \allocatedTasks} 
        \batteryVoltageN
        \cdot
        \parens{\batteryCurrent_{\taskIndex}^{(\robotIndex)}}^{\PeukertExponent}
        \cdot 
        \taskDurationK,
    \equlabel{total_battery_constraint}
\end{equation}
where $\batteryInitN$ is the robot's initial battery capacity, $\batteryInterN$ is the energy consumed during inter-task transitions, $\batteryVoltageN$ is its operating voltage, and $\taskDurationK$ is the duration of the $\taskIndex$-th task.

%% file: sections/5_approach.tex
\section{TRAITS: Optimization Framework}
\begin{figure}[t]
    \centering
    \includegraphics[width=\linewidth]{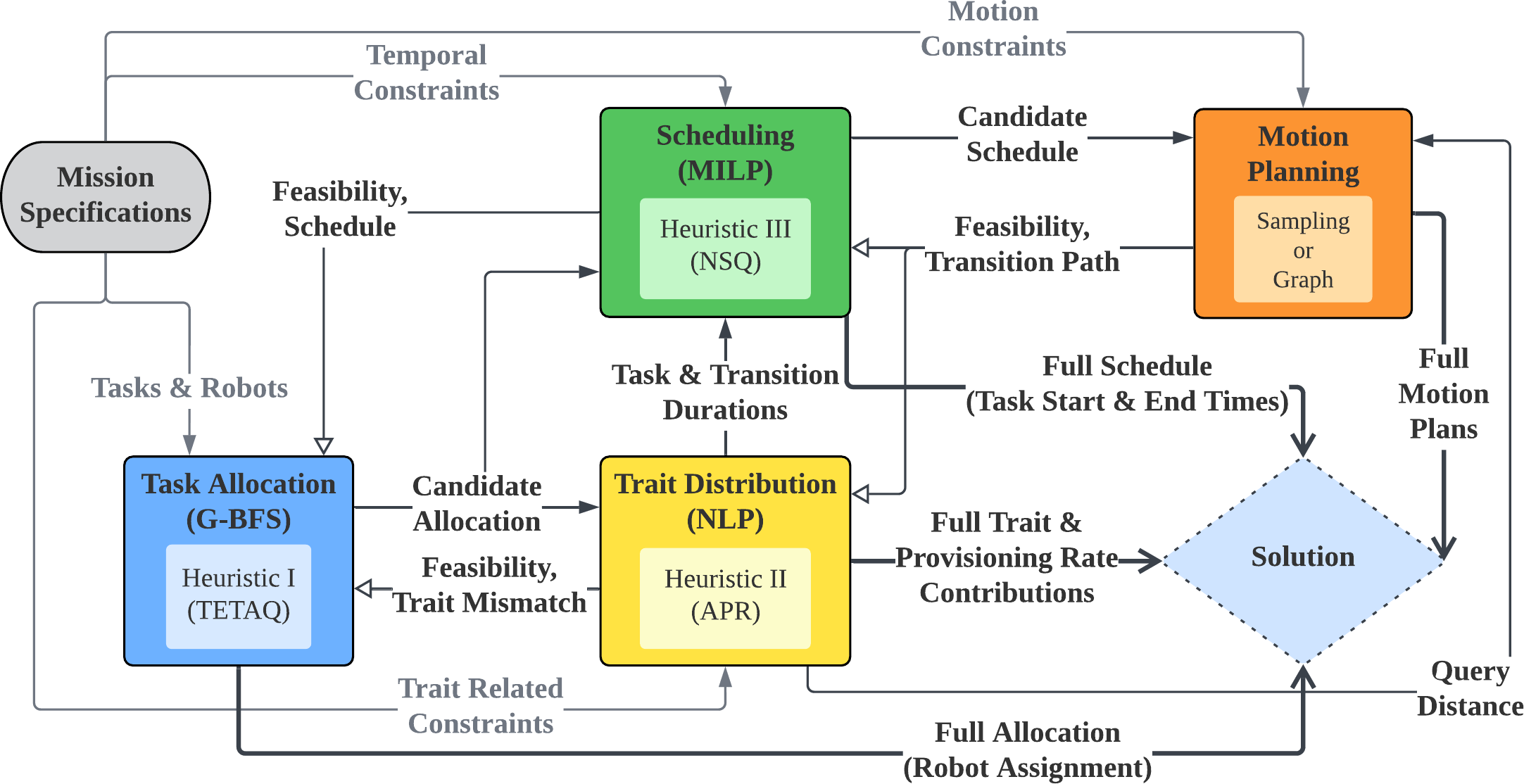}
    \caption{
        High-level architecture of the \acrshort*{traits} framework.
    }
    \Description{A figure shows the high-level overview of our TRAITS framework.}
    \figlabel{traits_framework}
\end{figure}
The high-level overview of the \acrshort*{traits} framework is shown in \figref{traits_framework} (refer \algoref{traits}), consisting of four different layers.
The \textit{task allocation} layer incrementally adds robots to tasks and conveys allocation to trait distribution and scheduling layers.
The \textit{trait distribution} layer computes feasible trait provisions under battery constraints and communicates the estimated task durations, obtained from the nonlinear program (NLP), to the scheduling layer.
The \textit{scheduling} layer queries the \textit{motion planning} layer for feasible paths and their corresponding transition durations, and computes an efficient makespan schedule using a mixed-integer linear program (MILP).
This section provides an in-depth examination of each layer and its components.
Collision avoidance is abstracted away in our framework and is assumed to be handled by a low-level control layer at deployment.
\input{pseudocode/algorithm}
\subsection{Heuristic-based Task Allocation}
The task allocation layer performs a greedy best-first search that incrementally assigns robots to tasks based on a heuristic evaluation.
The search employs the following three heuristics. 
The \acrfull*{apr} is defined as
\begin{equation}
    f_{apr} \parens{\traitMismatch, \traitRateMismatch} \triangleq
    \frac{ \gamma \cdot \traitMismatch}{\sum_k \sum_u \desiredTraitsKU} + \frac{(1 - \gamma) \traitRateMismatch}{\sum_k \sum_u \desiredTraitRatesKU} \; \in \ccint{0, 1},
    \equlabel{apr}
\end{equation}
where 
$\gamma \in \ooint{0, 1}$ is a user-defined hyperparameter that balances trait and trait-provisioning rate mismatches.
The variables $\traitMismatch$ and $\traitRateMismatch$ represent the total trait and trait-provisioning rate mismatches, respectively, and are defined as
\begin{align}
    %
    & \traitMismatch = \sum_{\taskIndex \in \taskIndexSet} \sum_{\traitIndex \in \traitIndexSet} \max (0, \underbrace{\desiredTraitsKU - \coalitionTraitKU}_{\traitMismatchKU}), & \equlabel{trait_mismatch} \\
    %
    %
    & \traitRateMismatch = \sum\limits_{\taskIndex \in \taskIndexSet} \sum\limits_{\traitIndex \in \gradualProvisioningTrait} \max  (0, \underbrace{\desiredTraitRatesKU - \coalitionTraitRateKU}_{\traitRateMismatchKU}). & \equlabel{trait_rate_mismatch}
\end{align}
A trait (provisioning rate) mismatch is defined as the non-negative shortfall between the required and provided trait (provisioning rate), with any surplus treated as zero mismatch.

The \acrfull*{nsq} is defined as
\begin{equation}
    f_{nsq}(\makespan) \triangleq \frac{\makespan- \makespanLB}{\makespanUB - \makespanLB} \; \in \ccint{0, 1},
    \equlabel{nsq}
\end{equation}
where $\makespan$ is the makespan of the given schedule, $\makespanLB = \sum\limits_{\taskIndex \in \taskIndexSet} \taskDurationKS$, and $\makespanUB =  \sum\limits_{\taskIndex \in \taskIndexSet} 
\sqBrackets{
\taskDurationKS + 
\max\limits_{\traitIndex \in \gradualProvisioningTrait} \parens{\frac{\desiredTraitsKU}{\desiredTraitRatesKU}} + 
\frac{2 \transitionLength_{\max}}{\velocity_{\min}}}$ 
where $\transitionLength_{\max}$ is the distance of the longest possible path in the map and $\velocity_{\min}$ is the slowest robot's speed.

The \acrfull*{tetaq} is the heuristic with user-defined hyperparameter $\alpha$, which is defined as
\begin{equation}
    f_{tetaq} \triangleq \alpha f_{apr} \parens{\traitMismatch, \traitRateMismatch} + (1 - \alpha) f_{nsq} \parens{\makespan} \; \in \ccint{0, 1}.
    \equlabel{tetaq}
\end{equation}
The task allocation layer returns a solution when allocation satisfies $f_{apr} = 0$ and $\makespan \leq \makespanUB$; otherwise, the search continues until a feasible solution is found or the user-defined timeout is reached.

\subsection{Incorporating Battery Consumption}
While many existing heterogeneous multi-robot task allocation approaches assume that robots always travel at their maximum speeds~\cite{neville2021interleaved, liu2026ijrr}, we relax this assumption by computing each robot's maximum transition speed based on its initial battery capacity.
The distance function $\transitionLength(\initialStateK[], \terminalStateK[], \robotIndex)$ returns the length of a traversable path from the initial position $\initialStateK[]$ to the terminal position $\terminalStateK[]$ for robot $\robotIndex$.

\subsubsection{Consumption during task execution}
This paper assumes the coalition moves at a uniform speed during the task execution. 
Accordingly, the intra-task transition speed $\coalitionIntraVelocityK$ for robots assigned to the $\taskIndex$-th task is obtained by maximizing it subject to~\constrref{intra_transition_velocity}.
We further assume that the path is constrained by the robot with the widest bounding radius, denoted as $\robotIndex_{wide}^{\taskIndex}$. 
Thus, the intra-task transition duration of the $\taskIndex$-th task is defined as
\begin{equation}
    \intraTransitionDuration = 
        \frac{\transitionLength 
        \parens{\initialStateK, \terminalStateK, 
        \robotIndex_{wide}^{\taskIndex}}}{\coalitionIntraVelocityK}.
    \equlabel{intra_task_transition_duration}
\end{equation}

\subsubsection{Consumption during task transitions}
The two inter-task transitions of concern are:
(i) from a robot's initial position to its first task $\initialTransitionDurationK$, and 
(ii) between consecutive tasks $\transitionDurationIJ$.
The average inter-task transition speed $\velocityInterN \in [0, \maxVelocityN]$ is the highest feasible speed under the battery constraint.
Robots may use all remaining energy after task execution, as shown in \constrref{total_battery_constraint}, as long as their speed does not exceed their maximum limit $\maxVelocityN$.
To this end, let $\batteryN_{\inter}$ denote the total battery energy consumed by $\robotIndex$-th robot during inter-task transitions over its schedule. Then
\begin{equation}
    \batteryN_{\inter} \geq 
            \batteryVoltageN 
            \cdot
            \parens{
                \batteryIdleCurrentN
                +
                \velocityCurrentCoeffN
                \cdot
                \velocityInterN
            }^{\PeukertExponent}
            \cdot
            \interTransitionDurationN,
    \equlabel{inter_transition_battery_constraint}
\end{equation}
where $\interTransitionDurationN=\frac{\interTransitionLengthN}{\velocityInterN}$ is the total inter-task transition duration of the $\robotIndex$-th robot.

Nonetheless, determining the exact value of $\interTransitionDurationN$ is computationally intractable, as the precise total traveling distance $\interTransitionLengthTrueN$ depends on task orderings---an NP-hard combinatorial problem.
Therefore, \acrshort*{traits} employs an overestimation of inter-task transition distance 
$\parens{\text{i.e.,} \; \interTransitionLengthTrueN \leq \interTransitionLengthEstimateN}$, where
    \begin{equation}
        \interTransitionLengthEstimateN = \sum\limits_{\taskIndex \in \allocatedTasks} \transitionLength
        \parens{\initialConfigurationsN, \initialStateK, \robotIndex}
        + \transitionLength
        \parens{\terminalStateK, \initialConfigurationsN, \robotIndex},
        \equlabel{inter_transition_ub}
    \end{equation}
\noindent
that is the sum of distances between the robot's initial position $\initialConfigurationsN$ and its assigned tasks' start $\initialStateK$ and end $\terminalStateK$ positions.
Once $\velocityInterN$ is obtained from \constrref{inter_transition_battery_constraint}, the initial transition duration for the $\taskIndex$-th task is calculated as
$
    \initialTransitionDurationK = \max\limits_{\robotIndex \in \allocatedRobots} \brackets{ 
        \frac{\transitionLength \left( \initialConfigurationsN, \initialStateK, \robotIndex \right)}
        { \velocityInterN} 
    }
$
while the inter-task transition duration between 
the $i$-th and $j$-th tasks is given by
$\transitionDurationIJ = \max\limits_{\robotIndex \in \allocatedRobots[i] \bigcap \allocatedRobots[j]} \frac{\transitionLength \left( \terminalStateK[i], \initialStateK[j], \robotIndex \right)}{\velocityInterN}$, effectively establishing the lower bounds on the time required for robot transitions.

\subsection{Temporal Constraints}
The temporal relationship between tasks can be modeled using synchronization and precedence (SP) constraints, as well as time window (TW) constraints~\cite{nunes2017taxonomy}.
This framework incorporates three forms of temporal constraints: precedence~\precedenceConstraints, mutex~\mutexConstraints, and deadlines \absoluteDeadlineConstraints, \relativeDeadlineConstraints.
A precedence constraint $\parens{\taskI[i] \prec \taskI[j]} \in \precedenceConstraints$ specifies that the $i$-th task must be completed before the $j$-th task begins.
A mutex constraint $\parens{\taskI[i] \leftrightarrow \taskI[j]} \in \mutexConstraints$, with $\precedenceConstraints \cap \mutexConstraints = \emptyset$, ensures that the two tasks cannot be executed concurrently.
An absolute deadline constraint $\parens{p < t_{abs}} \in \absoluteDeadlineConstraints$ requires the task's start or finish time $p \in \R_{\geq 0}$ to fall within a specified bound.
A relative deadline constraint $\parens{p_i, p_j, t_{rel}} \in \relativeDeadlineConstraints$ enforces that the time difference between two specified time points---$p_i$ of $i$-th task 
and $p_j$ of $j$-th task---must be less than $t_{rel}$, assuming $\taskI[i] \prec \taskI[j]$.
The values of $p_i$ and $p_j$ can represent either the start or end time of their respective tasks.

\subsection{Nonlinear Programming (NLP) Trait Distributor}
A key contribution of our work is that, unlike prior frameworks relying on simple matrix operations, \acrshort*{traits} formulates trait provisioning jointly per task and per robot, incorporating both provisioning rate and battery constraints into the allocation process.
The problem of determining robot-to-task trait provisions is formulated as a nonlinear program with hierarchical objectives and constraints.%
\input{equations/nlp_trait_distribution}
where the hierarchical objective $f_{obj}$ is to: 
\circled{1} minimize $f_{apr}$~in \equref{apr}, 
\circled{2} minimize task duration $\taskDurationK$, and 
\circled{3} maximize transitions velocities 
$\velocityInterN$.
This formulation facilitates the identification of feasible solutions that satisfy trait requirements while minimizing the makespan, subject to battery constraints.
Objectives with higher priority (i.e., those indicated by smaller numbers inside circles) must not be compromised to improve lower-priority objectives.

\subsubsection{Trait Provision}
Constraints \equnum{robot_trait_max_constraint}--\equnum{coalition_CT_trait_constraint} define the trait provision requirements.
\constrref{robot_trait_max_constraint} ensures that a robot's trait provision does not exceed its individual capacity, while
\equnum{robot_ET_trait_max_constraint} limits total use of exhaustible traits to the robot's initial levels.
Constraints \equnum{coalition_NCT_trait_constraint} and \equnum{coalition_CT_trait_constraint} enforce the coalition's trait provision for \nonCumulativeTrait~and \cumulativeTrait, respectively.

\subsubsection{Trait-Provisioning Rate}
Constraints \equnum{robot_trait_rate_max_constraint}--\equnum{coalition_GPT_trait_constraint} address trait-provisioning rates across assigned robots. 
\constrref{robot_trait_rate_max_constraint} enforces each robot's maximum rate limit, while 
\equnum{coalition_NCT_trait_rate_constraint} ensures that the team-level capability for an NCT is determined by the least capable robot in the team.
The coalition-level simultaneous provisioning rate model, consistent with \equref{trait_provision_rate_definition}, is enforced by
\equnum{coalition_GPT_trait_constraint}.

\subsubsection{Task Duration and Energy}
Constraints \equnum{nonGPT_trait_duration_constraint}--\equnum{task_duration_constraint} establish task durations based on trait provision and intra-transition durations, and \equnum{battery_constraint} ensures that energy consumption does not exceed the robot's initial battery capacity.

\subsection{Mixed-Integer Linear Programming (MILP) Scheduler}
Once the \acrshort*{nlp}-based trait distributor determines $\taskDurationK$ and $\velocityInterN$, 
the \acrshort*{milp} scheduler computes $\initialTransitionDurationK$ and $\transitionDurationIJ$ as part of the scheduling process
and searches for a schedule that minimizes the makespan $\makespan$ as follows:
\input{equations/milp_scheduler}

The makespan of the schedule is defined as the finish time of the last task \equnum{milp_last_task}.
Each task must finish at least $\taskDurationK$ after it begins execution \equnum{milp_task_finish_time}.
A task may commence only after the initial robot transitions have completed \equnum{milp_task_start_after_initial_transition}. 
If a precedence constraint applies, it may commence only after the transitions from its predecessor tasks have also completed \equnum{milp_task_start_precedence_constraint}.
The Boolean indicator $\mutexIndicatorIJ = 1$ \equnum{milp_mutex_indicator} encodes a mutex constraint reduced to $\taskI[i] \prec \taskI[j]$, and $\mutexIndicatorIJ = 0$ otherwise.
This introduces combinatorial complexity, making the scheduling problem an instance of an NP-hard class.
Finally, \equnum{milp_absolute_deadline} and \equnum{milp_relative_deadline} enforce absolute and relative deadline constraints, respectively.

\subsection{Solution}
A solution is considered feasible when the following three conditions are satisfied:
(i) the trait mismatch $\traitMismatch = 0$, (ii) the trait-provisioning rate mismatch $\traitRateMismatch = 0$, and (iii) the makespan $\makespan \leq \makespanUB$.
If these conditions are met, the solution is represented as the tuple $\traitsSolution$, where:
\circled{1} $\allocation\in \brackets{0, 1}^{\numTasks \times \numRobots}$ is the task allocation matrix, where $\allocationKN = 1$ if the $\robotIndex$-th robot is assigned to the $\taskIndex$-th task, and 0 otherwise.
\circled{2} $\traitProvisions$ and  $\traitProvisionRates$ denote the optimized trait values and provisioning rates, respectively, for every trait of each robot assigned to each task.
\circled{3} $\schedule$ is the task schedule, containing each task's start time $\taskStartTime[]$ and finish time $\taskEndTime[]$.
\circled{4} $\motionPlans$ contains the motion plans for all robots, including both inter- and intra-task transitions.

%% file: pseudocode/algorithm.tex
\begin{algorithm}[tbh]
\DontPrintSemicolon
    \small
    \KwIn{$ 
    \precedenceConstraints, \absoluteDeadlineConstraints, \relativeDeadlineConstraints,
    \desiredTraitsMatrix, \desiredTraitRatesMatrix, \robotTraitInitNU[][], \robotMaxTraitRateNU[][], \batteryInit, \initialConfigurations,
    \alpha, \gamma$}
    \KwOut{$\traitsSolution$}
    $pq \gets \text{PriorityQueue}\left(\{ root \}\right)$\; 
    \While{pq \textbf{not} empty} 
    {
        $node \gets pq.\text{pop}()$\;
        \If {$node.apr=0$ \textbf{and} $node.nsq \leq 1$}
        {
           \Return $\traitsSolution$\; 
        }
        \For {$\allocation \in \text{generateSuccessor}(node)$}
        {
            $\robottrait, \robottraitRate, \traitMismatch, \traitRateMismatch,
            \taskDuration, \velocityInter \gets \text{traitDistributor}\parens{\allocation}$ \;
           
            ${apr} \gets f_{apr}(\traitMismatch, \traitRateMismatch, \gamma)$ \;
            
            $\makespan, \schedule, \motionPlans \gets \text{scheduler}\parens{\allocation, \precedenceConstraints, \absoluteDeadlineConstraints, \relativeDeadlineConstraints,
            d, 
            \velocityInter}$ \;
            
            ${nsq} \gets f_{nsq}(\makespan)$ \;
            ${tetaq} \gets f_{tetaq}(\alpha, apr, nsq)$ \;
            $child \gets \angled{tetaq, apr, nsq, \allocation, \robottrait, \robottraitRate, \schedule, \motionPlans}$ \;
            $pq$.push($child$) \;
        }
    }
    \Return Null \;
    
\caption{TRAITS}
\algolabel{traits}
\end{algorithm}

%% file: equations/nlp_trait_distribution.tex
\begin{subequations}
    \begin{flalign}
        & \text{minimize} \; f_{obj} & \forall \taskIndex \in \taskIndexSet, \forall \robotIndex \in \robotIndexSet & \nonumber \\
        %
        %
        & \hspace{5pt} s.t. \; \robotTraitInitNU \geq \robottraitKNU & \forall \traitIndex \equlabel{robot_trait_max_constraint} \\
        & \hspace{19pt} \robotTraitInitNU \geq \sum_{\taskIndex \in \allocatedTasks} \robottraitKNU & \forall \traitIndex \in \exhaustibleTrait \equlabel{robot_ET_trait_max_constraint} \\
        & \hspace{19pt} \coalitionTraitKU \leq \robottraitKNU & \forall \traitIndex \in \nonCumulativeTrait \equlabel{coalition_NCT_trait_constraint} \\
        & \hspace{19pt}  \coalitionTraitKU = \sum\limits_{\robotIndex \in \allocatedRobots} \robottraitKNU & \forall \traitIndex \in \cumulativeTrait \equlabel{coalition_CT_trait_constraint} 
    \end{flalign}
    \begin{flalign}
        %
        %
        & \hspace{19pt}  \robotMaxTraitRateNU \geq \robottraitRateKNU & \forall \traitIndex \equlabel{robot_trait_rate_max_constraint} \\
        & \hspace{19pt}  \coalitionTraitRateKU \leq \robottraitRateKNU & \forall \traitIndex \in \nonCumulativeTrait \equlabel{coalition_NCT_trait_rate_constraint} \\
        & \hspace{19pt}  \coalitionTraitRateKU = \coalitionTraitKU / \taskDurationKUQ & \forall \traitIndex \in \gradualProvisioningTrait \equlabel{coalition_GPT_trait_constraint} \\ 
        %
        %
        & \hspace{19pt}  \taskDurationKNUQ = 0 & \forall  \traitIndex \notin \gradualProvisioningTrait \equlabel{nonGPT_trait_duration_constraint} \\
        & \hspace{19pt}  \taskDurationKNUQ = \robottraitKNU / \robottraitRateKNU & \forall  \traitIndex \in \gradualProvisioningTrait \equlabel{GPT_duration_constraint} \\
        & \hspace{19pt}  \taskDurationKUQ \geq \taskDurationKNUQ & \forall \traitIndex \equlabel{task_trait_duration_constraint} \\
        & \hspace{19pt}  \taskDurationK = \taskDurationKS + \taskDurationKQ + \intraTransitionDuration 
        \equlabel{task_duration_constraint} \\
        %
        %
        & \hspace{19pt}  \batteryN \leq \batteryInitN & \equlabel{battery_constraint}
    \end{flalign}
    \equlabel{nlp}
\end{subequations}

%% file: equations/milp_scheduler.tex
\begin{subequations}
\small
    \begin{flalign}
        & \text{minimize} \; \makespan & & \nonumber \\
        & \hspace{3mm} s.t. \; \makespan \geq \taskEndTime & \forall \taskIndex \in \taskIndexSet \equlabel{milp_last_task} \\
        & \hspace{23pt} \taskEndTime[\taskIndex] \geq \taskStartTime[\taskIndex] + \taskDurationI[\taskIndex] & \forall \taskIndex \in \taskIndexSet \equlabel{milp_task_finish_time} \\ 
        & \hspace{23pt} \taskStartTime[\taskIndex] \geq \initialTransitionDurationK[\taskIndex] & \forall \taskIndex \in \taskIndexSet \equlabel{milp_task_start_after_initial_transition} \\
        & \hspace{23pt}  \taskStartTime[j] \geq \taskEndTime[i] + \transitionDurationIJ & \forall (i, j) \in \precedenceConstraints \equlabel{milp_task_start_precedence_constraint} \\
        & \hspace{23pt}  \taskStartTime[j] \geq \taskEndTime[i] + \transitionDurationIJ - M (1 - \mutexIndicatorIJ) & \forall (i, j) \in \mutexConstraints \equlabel{milp_task_start_mutex_ij} \\
        & \hspace{23pt}  \taskStartTime[i] \geq \taskEndTime[j] + \transitionDurationJI - M \mutexIndicatorIJ & \forall (i, j) \in \mutexConstraints \equlabel{milp_task_start_mutext_ji} \\
        & \hspace{23pt}  \mutexIndicatorIJ \in \{0, 1\} & \forall (i, j) \in \mutexConstraints \equlabel{milp_mutex_indicator} \\
        & \hspace{23pt}  p \leq t_{abs} & \forall \left( p, t_{abs} \right) \in T_{abs} \equlabel{milp_absolute_deadline} \\
        & \hspace{23pt}  p_j - p_i \leq t_{rel} & \forall \left( p_i, p_j, t_{rel} \right) \in T_{rel} \equlabel{milp_relative_deadline}
    \end{flalign}
    \equlabel{milp}
\end{subequations}

%% file: sections/6_evaluation.tex
\section{Evaluation}
We evaluate proposed framework across three different aspects.
First, we compare \acrshort*{traits} against two state-of-the-art trait-based \acrshort*{mrta} frameworks, \acrshort*{itags}~\cite{neville2021interleaved} and \acrshort*{ctas}~\cite{fu2023robust} (see \tabref{framework comparison}). 
Second, we conduct stress tests by varying the number of tasks $\numTasks$ and robots $\numRobots$. 
Third, we analyze the impact of hyperparameters on search performance. 
The evaluation is based on 400 randomized experiments across various scenarios in a simulated warehouse environment with diverse trait requirements. 
In each trial, parameters are sampled as follows: $\numRobots \in \ccint{5, 30}$, $\numTasks \in \ccint{5, 40}$, and $\PeukertExponent \in [1.01, 1.15]$.
Additionally, $\robotTraitInitNU, \robotMaxTraitRateNU, \batteryInitN$ are sampled $\ocint{0, 1}$ relative to each robot's respective maximum values.
Each experiment includes at least one instance of precedence $\precedenceConstraints$, absolute deadline $\absoluteDeadlineConstraints$, and relative deadline $\relativeDeadlineConstraints$ constraints.
\begin{table}[tb]
    \caption{
        Comparison of \acrshort*{traits}, \acrshort*{itags}, and \acrshort*{ctas} across key attributes.
        While uncertainty is not the focus of this work, it is considered exclusively in CTAS.
    }
    \tablabel{framework comparison}
    \resizebox{\linewidth}{!}{
        \begin{tabular}{lccc}
            \toprule
              Considerations & \multicolumn{1}{c}{\acrshort*{traits} (ours)} & \multicolumn{1}{c}{\acrshort*{itags}~\cite{neville2021interleaved}} & \multicolumn{1}{c}{\acrshort*{ctas}~\cite{fu2023robust}}\\
             \midrule
             Uncertainty control & \xmark & \xmark & \cmark \\
             Motion planning & \cmark & \cmark & \xmark \\
             Time-varying trait & \cmark & \xmark & \xmark \\
             Trait-provisioning rate & \cmark & \xmark & \xmark \\
             Variable task duration & \cmark & \xmark & \xmark \\
             Battery depletion & \cmark & \xmark & \xmark \\
             Deadline constraint & \cmark & \xmark & \xmark \\
             \bottomrule
        \end{tabular}
    }
\end{table}
All experiments were designed such that robots possess sufficient traits and battery to handle assigned tasks while forming meaningful coalitions. 
Each scenario includes both exhaustible and gradual-provisioning traits, and each task requires at least one cumulative or one non-cumulative trait. 
To observe the hyperparameter effects, the robots were diversified to include low trait-provisioning rate with large capacity, standard, and high trait-provisioning rate with limited capacity. 
Battery current coefficients were modeled using technical specifications from Clearpath Robotics' Husky~\cite{clearpath_husky_specs}.
It is often difficult to even establish the existence of a solution.
To address this, we incorporated a necessary condition into the framework: the initial set of robot teams must collectively possess sufficient trait to cover the requirement of the tasks.
If this condition is not satisfied at the outset, the problem is immediately declared infeasible.
\acrshort*{traits} framework leverages off-the-shelf libraries for motion planning~\cite{sucan2012the-open-motion-planning-library} and optimizations~\cite{gurobi}.
All experiments were conducted on an AMD Ryzen 5950X CPU and 128 GB of RAM.

\subsection{Comparison with Baselines}
\begin{table}[bt]
    \caption{
        Performance comparison of \acrshort*{traits}, \acrshort*{itags}, and \acrshort*{ctas}.
        Values are $\mu~(\pm \sigma)$. 
        \textbf{Bold} indicates the best performance, which improves in the direction of the arrow.
        Results are based on 300 experiments with $\ccint{5, 25}$ tasks and $\ccint{5, 25}$ robots.
    }
    \tablabel{itags_benchmark}
    \resizebox{\linewidth}{!}{
        \begin{tabular}{lrrr}
            \toprule
              & \multicolumn{1}{c}{\acrshort*{traits} (ours)} & \multicolumn{1}{c}{\acrshort*{itags}~\cite{neville2021interleaved}} & \multicolumn{1}{c}{\acrshort*{ctas}~\cite{fu2023robust}}\\
             \midrule
             Plan feasibility [\%] $(\uparrow)$ & $\mathbf{100.0} \, (\pm \mathbf{0.0})$ & $51.7 \, (\pm 17.8)$ &  $63.6 \, (\pm 16.4)$ \\
             Task trait insufficiency [\%] $(\downarrow)$ & $\mathbf{0.0} \, (\pm \mathbf{0.0})$ & $11.5 \, (\pm 17.9)$ & $0.2 \, (\pm 2.0)$ \\
             Provisioning rate insufficiency [\%] $(\downarrow)$ & $\mathbf{0.0} \, (\pm \mathbf{0.0})$ & $41.6 \, (\pm 14.3)$ & $36.2 \, (\pm 16.5)$ \\
             Under-resourced robots [\%] $(\downarrow)$ & $\mathbf{0.0} \, (\pm \mathbf{0.0})$ & $26.7 \, (\pm 13.7)$ & $14.3 \, (\pm 9.3)$ \\
             C-rating violations [\%] $(\downarrow)$ & $\mathbf{0.0} \, (\pm \mathbf{0.0})$ & $47.5 \, (\pm 14.9)$ & $37.7 \, (\pm 16.4)$ \\
             Battery-capacity violations [\%] $(\downarrow)$ & $\mathbf{0.0} \, (\pm \mathbf{0.0})$ & $29.7 \, (\pm 10.1)$ & $23.9 \, (\pm 11.1)$ \\
             Deadline violations [\%] $(\downarrow)$ & $\mathbf{0.0} \, (\pm \mathbf{0.0})$ & $5.0 \, (\pm 12.8)$ & $22.3 \, (\pm 18.9)$ \\
             Computation time [s] $(\downarrow)$ & $200.77 (\pm 253.26)$ & $\mathbf{31.12} (\pm \mathbf{27.75})$ & $207.81 (\pm 258.32)$\\
             \bottomrule
        \end{tabular}
    }
\end{table}

We evaluated \acrshort*{traits} against two state-of-the-art trait-based \acrshort*{mrta} frameworks \acrshort*{ctas}~\cite{fu2023robust} and \acrshort*{itags}~\cite{neville2021interleaved}.
Specifically, we chose CTAS-O, a deterministic variant of the framework proposed in~\cite{fu2023robust}.
As shown in \tabref{itags_benchmark} and \figref{under_resourced_fraction}, \acrshort*{traits} outperforms both \acrshort*{ctas} and \acrshort*{itags} across multiple metrics, largely because they do not model trait exhaustibility, provisioning rates, battery constraints, and deadline constraints.
Since \acrshort*{ctas} and \acrshort*{itags} lack trait-pro\-visioning rates, their trait-provisioning durations are computed as 
$\taskDurationKQ = \max\limits_{\traitIndex \in \gradualProvisioningTrait} \parens{\frac{\desiredTraitsKU}{\desiredTraitRatesKU}}$.
Moreover, CTAS is an anytime algorithm, meaning its solution quality improves as computation time increases. To accommodate this, the experiment was structured so that CTAS returns a solution upon finding the optimal one, returns the best solution found at the \acrshort*{traits} timeout if a feasible one exists, or continues running until a solution is found.

\acrshort*{traits} incurs greater computation time due to its modeling of rate-dependent trait provision, deadlines, and battery-aware scheduling.
\acrshort*{itags} and \acrshort*{ctas} achieve 51.7\% and 63.6\% plan feasibility, respectively, often generating solutions that \acrshort*{traits} rejects due to constraint violations---such as trait and provisioning rate insufficiencies, under-resourced robots, C-rating violations, and battery-capacity violations---which are rigorously handled by our proposed \acrshort*{traits} framework.
\figref{under_resourced_fraction} corroborates that \acrshort*{traits} consistently avoids overcommitment by ensuring that robots have sufficient traits to execute their assigned tasks, unlike \acrshort*{itags} and \acrshort*{ctas}, which tend to assign tasks beyond the robots' capabilities.

\begin{figure*}[t]
    \centering
    \includegraphics[width=0.8\linewidth]{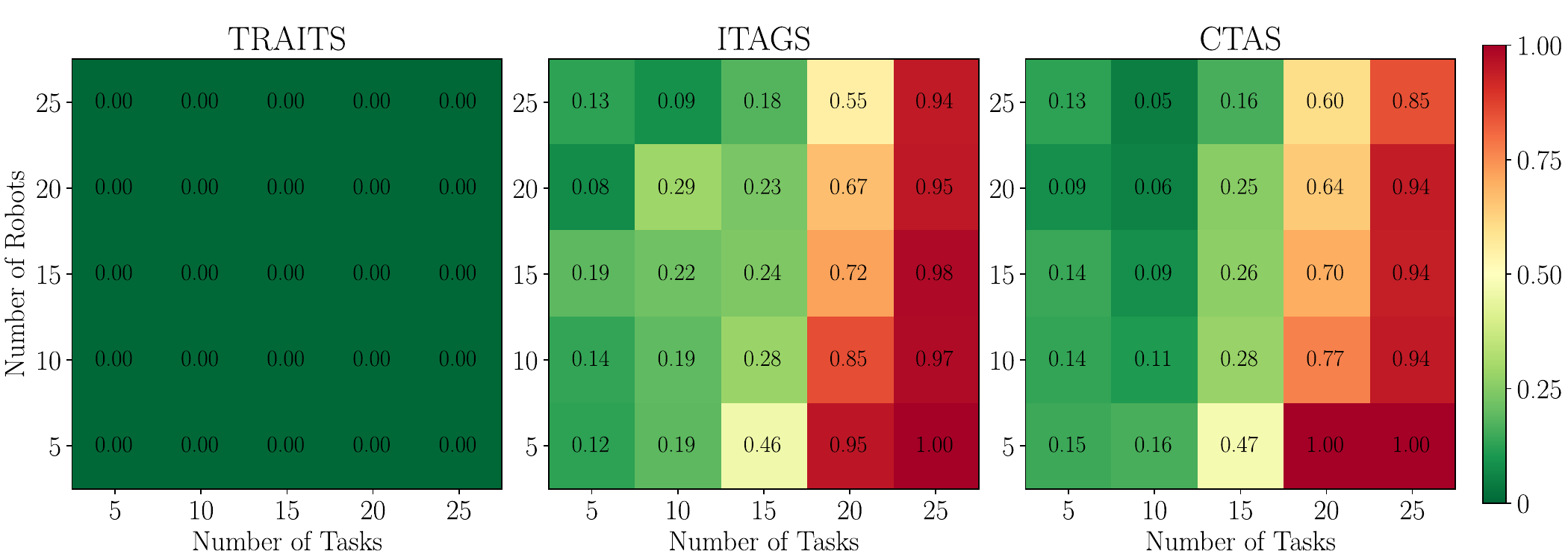}
    \caption{
        Fraction of under-resourced robots (robots assigned to tasks without sufficient traits).
        Lower values (\textcolor{ForestGreen}{green}) indicate better performance.
        \acrshort*{traits} consistently avoids assigning robots to tasks that exceed their trait capacities, whereas \acrshort*{itags} and \acrshort*{ctas} tend to assign insufficiently resourced robots as the task load grows for a fixed team size.
        This behavior partly stems from the assumption of irreducible traits in \acrshort*{itags} and \acrshort*{ctas}.
        Under-resourcing declines with increasing team size.
    }
    \figlabel{under_resourced_fraction}
    \Description{figure shows the fraction of under-resourced robots across various combination of tasks and robots for different frameworks.
    }
\end{figure*}

\subsection{Scalability and Hyperparameter Sensitivity}
Stress tests were conducted to analyze computation time as a function of the number of tasks $\numTasks$ and robots $\numRobots$.
As shown in \figref{stress_test}, the number of tasks has a significantly greater impact on computation time than the number of robots, with differences reaching up to an order of magnitude.
In general, computation time increases as both $\numTasks$ and $\numRobots$ grow, primarily due to the expansion in the number of variables in the trait distribution and scheduling modules.
Owing to its NP-hard combinatorial structure, computational effort grows rapidly with the instance size, making increased runtime unavoidable as the problem scales.
Furthermore, when the task-to-robot ratio is high---meaning each robot is assigned to more than one task---computation time further increases.
This is attributed to the scheduling layer, which requires more time to resolve mutex constraints under high task-load conditions.

\begin{figure}[tb]
    \centering
  \begin{minipage}[b]{0.49\linewidth}
    \centering
    \includegraphics[width=\linewidth]{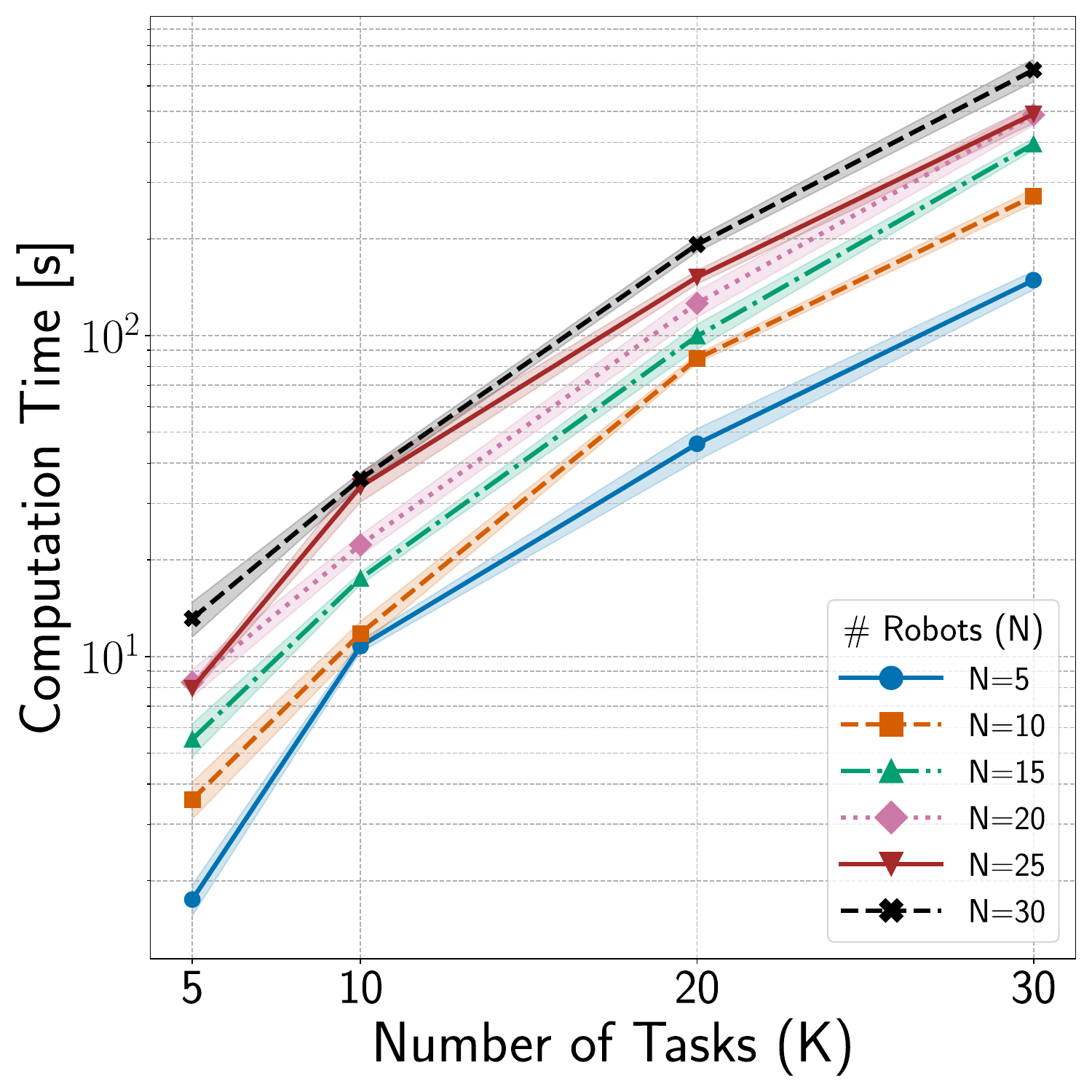}
  \end{minipage}
  \hfill
  \begin{minipage}[b]{0.49\linewidth}
    \centering
    \includegraphics[width=\linewidth]{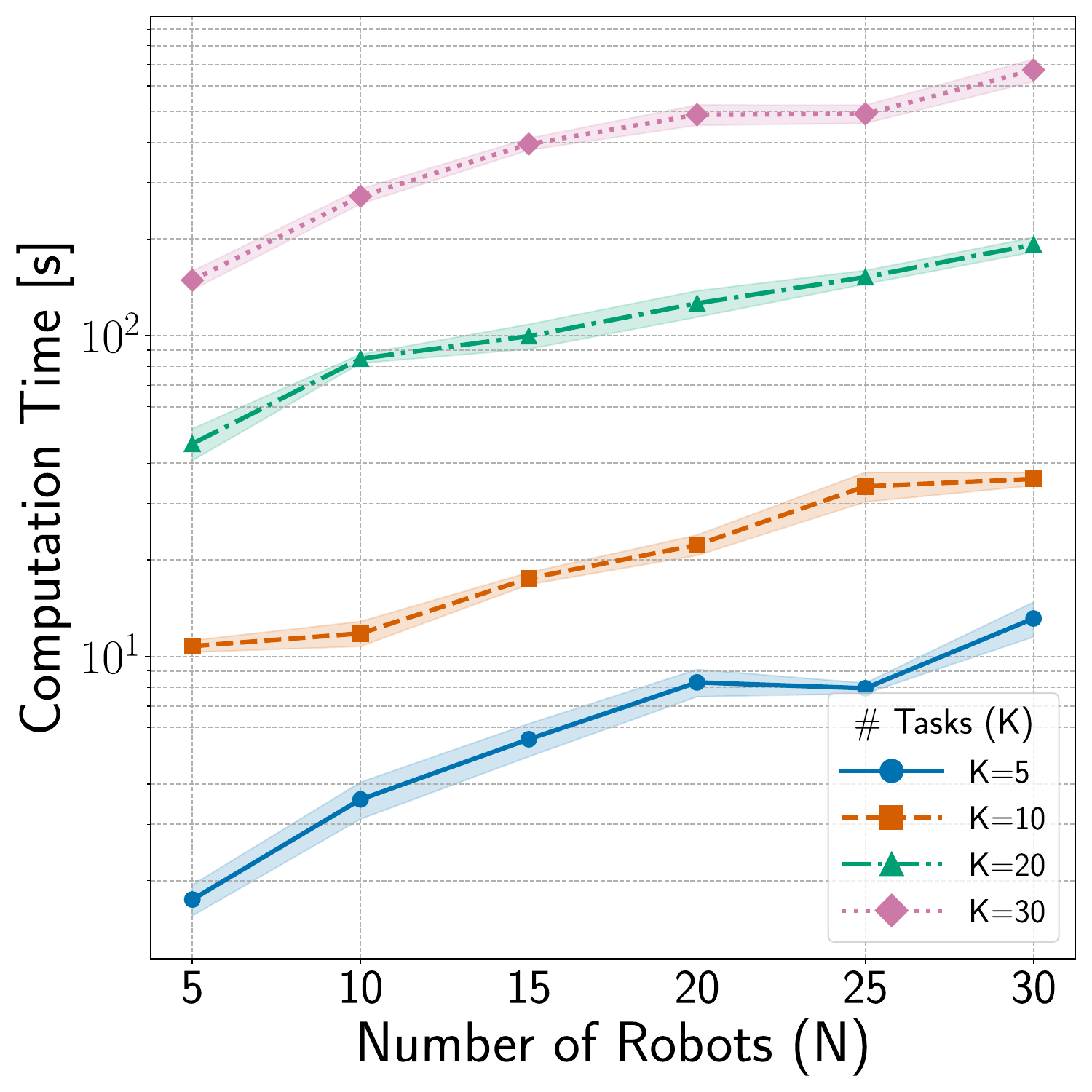}
  \end{minipage}
    \caption{
        Computation time as a function of the number of tasks $\numTasks$ (left) and the number of robots $\numRobots$ (right). 
        Solid dots indicate the mean, and the shaded regions represent the standard error computed from 120 experiments. 
        The results highlight that computation time increases with both parameters, with task count having a more pronounced impact.
    }
    \figlabel{stress_test}
    \Description{Stress tests result figures.}
\end{figure}

In this experiment, we also observed that the hyperparameters $\alpha$ and $\gamma$ both affect the makespan and computation times, but in distinct ways. 
$\alpha$ influences total task duration by favoring faster-provisioning robots as $\alpha \rightarrow 0$, which shortens task execution times.
It also significantly impacts computation time: smaller $\alpha$ values broaden the search over robot coalitions, increasing runtimes, while larger values lead to greedier solutions with shorter computation times by prioritizing feasibility over optimality---a trade-off noted in prior work~\cite{neville2021interleaved}. 
$\gamma$ affects transition durations by favoring robots closer to the tasks as $\gamma \rightarrow 0$ and reduces trait-provisioning durations as $\gamma$ increases. 
Although generally less impactful on computation time than $\alpha$, extreme $\gamma$ values can increase runtimes due to an imbalance in satisfying the trait mismatch $\traitMismatch$ and the trait-provisioning rate mismatch $\traitRateMismatch$.
If one is satisfied significantly earlier than the other, additional robot assignments may be needed, increasing search effort.

%% file: sections/7_conclusion.tex
\section{Conclusion}
This paper presents an offline heterogeneous time-extended multi-robot task allocation and scheduling framework that addresses the provision of exhaustible traits under battery constraints. 
We introduce the notions of trait provision and exhaustibility, which capture the temporal dynamics of traits and enable more expressive categorization of traits. 
Experimental results demonstrate our framework’s enhanced comprehensiveness and superior performance compared to state-of-the-art baselines, while also analyzing trade-offs between search efficiency and solution quality as influenced by two hyperparameters. 
Despite these strengths, our framework assumes constant trait-provisioning rates and presumes that paths remain collision-free regardless of the number of robots---limitations we aim to address in future work.
Overall, the proposed approach effectively manages a broader range of traits while maintaining computational tractability.

%% file: references.bib
@INPROCEEDINGS{aswale2023heterogeneous,
  author={Aswale, Ashay and Pinciroli, Carlo},
  booktitle={2023 IEEE/RSJ International Conference on Intelligent Robots and Systems (IROS)}, 
  title={Heterogeneous Coalition Formation and Scheduling with Multi-Skilled Robots}, 
  year={2023},
  volume={},
  number={},
  pages={5402-5409}
}

@article{banfi2022hierarchical,
  title={Hierarchical planning for heterogeneous multi-robot routing problems via learned subteam performance},
  author={Banfi, Jacopo and Messing, Andrew and Kroninger, Christopher and Stump, Ethan and Hutchinson, Seth and Roy, Nicholas},
  journal={IEEE Robotics and Automation Letters},
  volume={7},
  number={2},
  pages={4464--4471},
  year={2022},
  publisher={IEEE}
}

@InProceedings{beynier2007decentralized,
author="Beynier, Aur{\'e}lie
and Mouaddib, Abdel-Illah",
editor="Alami, Rachid
and Chatila, Raja
and Asama, Hajime",
title="Decentralized Markov Decision Processes for Handling Temporal and Resource constraints in a Multiple Robot System",
booktitle="Distributed Autonomous Robotic Systems 6",
year="2007",
publisher="Springer Japan",
address="Tokyo",
pages="191--200",
}

@inproceedings{calvo2024optimal,
  author={Calvo, Álvaro and Capitán, Jesús},
  booktitle={2024 IEEE International Conference on Robotics and Automation (ICRA)}, 
  title={Optimal Task Allocation for Heterogeneous Multi-robot Teams with Battery Constraints}, 
  year={2024},
  volume={},
  number={},
  pages={7243-7249},
  doi={10.1109/ICRA57147.2024.10611147}
}

@article{chen2011resource,
  title={Resource constrained multirobot task allocation based on leader--follower coalition methodology},
  author={Chen, Jian and Sun, Dong},
  journal={The International Journal of Robotics Research},
  volume={30},
  number={12},
  pages={1423--1434},
  year={2011},
  publisher={SAGE Publications Sage UK: London, England}
}

@INPROCEEDINGS{chen2023game,
  author={Chen, Shengkang and Lin, Tony X. and Zhang, Fumin},
  booktitle={2023 IEEE/RSJ International Conference on Intelligent Robots and Systems (IROS)}, 
  title={Game-Theoretical Approach to Multi-Robot Task Allocation Using a Bio-Inspired Optimization Strategy}, 
  year={2023},
  volume={},
  number={},
  pages={5388-5393},
  doi={10.1109/IROS55552.2023.10341947}}

@article{choi2009consensus,
  author={Choi, Han-Lim and Brunet, Luc and How, Jonathan P.},
  journal={IEEE Transactions on Robotics}, 
  title={Consensus-Based Decentralized Auctions for Robust Task Allocation}, 
  year={2009},
  volume={25},
  number={4},
  pages={912-926},
  doi={10.1109/TRO.2009.2022423}
}

@inproceedings{choi2010decentralized,
  title={Decentralized task allocation for heterogeneous teams with cooperation constraints},
  author={Choi, Han-Lim and Whitten, Andrew K and How, Jonathan P},
  booktitle={Proceedings of the 2010 American Control Conference},
  pages={3057--3062},
  year={2010},
  organization={IEEE}
}

@article{choudhury2022dynamic,
  title={Dynamic multi-robot task allocation under uncertainty and temporal constraints},
  author={Choudhury, Shushman and Gupta, Jayesh K and Kochenderfer, Mykel J and Sadigh, Dorsa and Bohg, Jeannette},
  journal={Autonomous Robots},
  volume={46},
  number={1},
  pages={231--247},
  year={2022},
  publisher={Springer}
}

@online{clearpath_husky_specs,
  author       = {{Clearpath Robotics}},
  title        = {Husky UGV Spec Comparison},
  year         = {2025},
  url          = {https://clearpathrobotics.com/husky-spec-comparison/},
}

@inproceedings{dias2004robust,
  author={Bernardine Dias, M. and Zinck, M. and Zlot, R. and Stentz, A.},
  booktitle={IEEE International Conference on Robotics and Automation, 2004. Proceedings. ICRA '04. 2004}, 
  title={Robust multirobot coordination in dynamic environments}, 
  year={2004},
  volume={4},
  number={},
  pages={3435-3442 Vol.4}}

@INPROCEEDINGS{dressler2005energy,
  author={Dressler, F. and Fuchs, G.},
  booktitle={Proceedings of the Fifth International Workshop on Robot Motion and Control, 2005. RoMoCo '05.}, 
  title={Energy-aware operation and task allocation of autonomous robots}, 
  year={2005},
  volume={},
  number={},
  pages={163-168},
  doi={10.1109/ROMOCO.2005.201418}}

@article{fu2023robust,
  author={Fu, Bo and Smith, William and Rizzo, Denise M. and Castanier, Matthew and Ghaffari, Maani and Barton, Kira},
  journal={IEEE Transactions on Robotics}, 
  title={Robust Task Scheduling for Heterogeneous Robot Teams Under Capability Uncertainty}, 
  year={2023},
  volume={39},
  number={2},
  pages={1087-1105},
  doi={10.1109/TRO.2022.3216068}
}

@article{gerkey2004formal,
  title={A formal analysis and taxonomy of task allocation in multi-robot systems},
  author={Gerkey, Brian P and Matari{\'c}, Maja J},
  journal={The International journal of robotics research},
  volume={23},
  number={9},
  pages={939--954},
  year={2004},
  publisher={SAGE Publications}
}

@ARTICLE{gombolay2018tercio,
  author={Gombolay, Matthew C. and Wilcox, Ronald J. and Shah, Julie A.},
  journal={IEEE Transactions on Robotics}, 
  title={Fast Scheduling of Robot Teams Performing Tasks With Temporospatial Constraints}, 
  year={2018},
  volume={34},
  number={1},
  pages={220-239}
}

@INPROCEEDINGS{gosrich2023multi,
  author={Gosrich, Walker and Mayya, Siddharth and Narayan, Saaketh and Malencia, Matthew and Agarwal, Saurav and Kumar, Vijay},
  booktitle={2023 IEEE International Conference on Robotics and Automation (ICRA)}, 
  title={Multi-Robot Coordination and Cooperation with Task Precedence Relationships}, 
  year={2023},
  volume={},
  number={},
  pages={5800-5806},
  doi={10.1109/ICRA48891.2023.10160998}}

@misc{gurobi,
  author = {{Gurobi Optimization, LLC}},
  title = {{Gurobi Optimizer Reference Manual}},
  year = 2025,
  url = "https://www.gurobi.com"
}

@article{ito2025energy,
  title={Energy-Aware Task Allocation for Teams of Multi-mode Robots},
  author={Ito, Takumi and Funada, Riku and Sampei, Mitsuji and Notomista, Gennaro},
  journal={arXiv preprint arXiv:2503.12787},
  year={2025}
}

@INPROCEEDINGS{kaminka2010adaptive,
  author={Kaminka, Gal A. and Erusalimchik, Dan and Kraus, Sarit},
  booktitle={2010 IEEE International Conference on Robotics and Automation}, 
  title={Adaptive multi-robot coordination: A game-theoretic perspective}, 
  year={2010},
  volume={},
  number={},
  pages={328-334}}

@inproceedings{koenig2006power,
  title={The power of sequential single-item auctions for agent coordination},
  author={Koenig, Sven and Tovey, Craig and Lagoudakis, Michail and Markakis, Vangelis and Kempe, David and Keskinocak, Pinar and Kleywegt, Anton and Meyerson, Adam and Jain, Sonal},
  booktitle={proceedings of the 21st national conference on Artificial intelligence-Volume 2},
  pages={1625--1629},
  year={2006}
}

@article{korsah2013comprehensive,
  title={A comprehensive taxonomy for multi-robot task allocation},
  author={Korsah, G Ayorkor and Stentz, Anthony and Dias, M Bernardine},
  journal={The International Journal of Robotics Research},
  volume={32},
  number={12},
  pages={1495--1512},
  year={2013},
  publisher={SAGE Publications Sage UK: London, England}
}

@article{leahy2022scratches,
  author={Leahy, Kevin and Serlin, Zachary and Vasile, Cristian-Ioan and Schoer, Andrew and Jones, Austin M. and Tron, Roberto and Belta, Calin},
  journal={IEEE Transactions on Robotics}, 
  title={Scalable and Robust Algorithms for Task-Based Coordination From High-Level Specifications (ScRATCHeS)}, 
  year={2022},
  volume={38},
  number={4},
  pages={2516-2535},
}

@article{lee2018resource,
  title={Resource-based task allocation for multi-robot systems},
  author={Lee, Dong-Hyun},
  journal={Robotics and Autonomous Systems},
  volume={103},
  pages={151--161},
  year={2018},
  publisher={Elsevier}
}

@misc{liu2026ijrr,
  title={Learning and Optimizing the Efficacy of Spatio-Temporal Task Allocation under Temporal and Resource Constraints}, 
  author={Jiazhen Liu and Glen Neville and Jinwoo Park and Sonia Chernova and Harish Ravichandar},
  year={2026},
  eprint={2601.02505},
  archivePrefix={arXiv},
  primaryClass={cs.RO},
  url={https://arxiv.org/abs/2601.02505}, 
}

@ARTICLE{mayya2021resilient,
  author={Mayya, Siddharth and D’antonio, Diego S. and Saldaña, David and Kumar, Vijay},
  journal={IEEE Robotics and Automation Letters}, 
  title={Resilient Task Allocation in Heterogeneous Multi-Robot Systems}, 
  year={2021},
  volume={6},
  number={2},
  pages={1327-1334},
}

@INPROCEEDINGS{mei2005energy,
  author={Yongguo Mei and Yung-Hsiang Lu and Hu, Y.C. and Lee, C.S.G.},
  booktitle={ICAR '05. Proceedings., 12th International Conference on Advanced Robotics, 2005.}, 
  title={A case study of mobile robot's energy consumption and conservation techniques}, 
  year={2005},
  volume={},
  number={},
  pages={492-497},
  doi={10.1109/ICAR.2005.1507454}}

@article{messing2022grstaps,
  title={Grstaps: Graphically recursive simultaneous task allocation, planning, and scheduling},
  author={Messing, Andrew and Neville, Glen and Chernova, Sonia and Hutchinson, Seth and Ravichandar, Harish},
  journal={The International Journal of Robotics Research},
  volume={41},
  number={2},
  pages={232--256},
  year={2022},
  publisher={SAGE Publications Sage UK: London, England}
}

@inproceedings{messing2023sampling,
  title={A Sampling-Based Approach for Heterogeneous Coalition Scheduling with Temporal Uncertainty},
  author={Messing, Andrew and Banfi, Jacopo and Stadler, Martina and Stump, Ethan and Ravichandar, Harish and Roy, Nicholas and Hutchinson, Seth},
  booktitle={Robotics: Science and Systems},
  year={2023}
}

@inproceedings{neville2020approximated,
  title={Approximated dynamic trait models for heterogeneous multi-robot teams},
  author={Neville, Glen and Ravichandar, Harish and Shaw, Kenneth and Chernova, Sonia},
  booktitle={2020 IEEE/RSJ International Conference on Intelligent Robots and Systems (IROS)},
  pages={7978--7984},
  year={2020},
  organization={IEEE}
}

@inproceedings{neville2021interleaved,
  author={Neville, Glen and Messing, Andrew and Ravichandar, Harish and Hutchinson, Seth and Chernova, Sonia},
  booktitle={2021 IEEE/RSJ International Conference on Intelligent Robots and Systems (IROS)}, 
  title={An Interleaved Approach to Trait-Based Task Allocation and Scheduling}, 
  year={2021},
  volume={},
  number={},
  pages={1507-1514},
  doi={10.1109/IROS51168.2021.9636569}
}

@article{neville2023ditags,
  title={D-ITAGS: a dynamic interleaved approach to resilient task allocation, scheduling, and motion planning},
  author={Neville, Glen and Chernova, Sonia and Ravichandar, Harish},
  journal={IEEE Robotics and Automation Letters},
  volume={8},
  number={2},
  pages={1037--1044},
  year={2023},
  publisher={IEEE}
}

@article{notomista2022resilient,
  author={Notomista, Gennaro and Mayya, Siddharth and Emam, Yousef and Kroninger, Christopher and Bohannon, Addison and Hutchinson, Seth and Egerstedt, Magnus},
  journal={IEEE Transactions on Robotics}, 
  title={A Resilient and Energy-Aware Task Allocation Framework for Heterogeneous Multirobot Systems}, 
  year={2022},
  volume={38},
  number={1},
  pages={159-179}
}

@article{nunes2017taxonomy,
    title = {A taxonomy for task allocation problems with temporal and ordering constraints},
    journal = {Robotics and Autonomous Systems},
    volume = {90},
    pages = {55-70},
    year = {2017},
    note = {Special Issue on New Research Frontiers for Intelligent Autonomous Systems},
    issn = {0921-8890},
    doi = {https://doi.org/10.1016/j.robot.2016.10.008},
    url = {https://www.sciencedirect.com/science/article/pii/S0921889016306157},
    author = {Ernesto Nunes and Marie Manner and Hakim Mitiche and Maria Gini}
}

@INPROCEEDINGS{park2023ritags,
  author={Park, Jinwoo and Messing, Andrew and Ravichandar, Harish and Hutchinson, Seth},
  booktitle={2023 IEEE/RSJ International Conference on Intelligent Robots and Systems (IROS)}, 
  title={Risk-Tolerant Task Allocation and Scheduling in Heterogeneous Multi-Robot Teams}, 
  year={2023},
  volume={},
  number={},
  pages={5372-5379},
  doi={10.1109/IROS55552.2023.10341837}}

@article{prorok2017impact,
  author={Prorok, Amanda and Hsieh, M. Ani and Kumar, Vijay},
  journal={IEEE Transactions on Robotics}, 
  title={The Impact of Diversity on Optimal Control Policies for Heterogeneous Robot Swarms}, 
  year={2017},
  volume={33},
  number={2},
  pages={346-358},
  doi={10.1109/TRO.2016.2631593}}

@article{ravichandar2020strata,
  title={STRATA: unified framework for task assignments in large teams of heterogeneous agents},
  author={Ravichandar, Harish and Shaw, Kenneth and Chernova, Sonia},
  journal={Autonomous Agents and Multi-Agent Systems},
  volume={34},
  pages={1--25},
  year={2020},
  publisher={Springer}
}

@inproceedings{schillinger2017multi,
  title={Multi-objective search for optimal multi-robot planning with finite ltl specifications and resource constraints},
  author={Schillinger, Philipp and B{\"u}rger, Mathias and Dimarogonas, Dimos V},
  booktitle={2017 IEEE International Conference on Robotics and Automation (ICRA)},
  pages={768--774},
  year={2017},
  organization={IEEE}
}

@article{shehory1998methods,
  title={Methods for task allocation via agent coalition formation},
  author={Shehory, Onn and Kraus, Sarit},
  journal={Artificial intelligence},
  volume={101},
  number={1-2},
  pages={165--200},
  year={1998},
  publisher={Elsevier}
}

@INPROCEEDINGS{shiroma2009comutar,
  author={Shiroma, Pedro M. and Campos, Mario F. M.},
  booktitle={2009 IEEE/RSJ International Conference on Intelligent Robots and Systems}, 
  title={CoMutaR: A framework for multi-robot coordination and task allocation}, 
  year={2009},
  volume={},
  number={},
  pages={4817-4824}}

@ARTICLE{shorinwa2023distributed,
  author={Shorinwa, Ola and Haksar, Ravi N. and Washington, Patrick and Schwager, Mac},
  journal={IEEE Transactions on Robotics}, 
  title={Distributed Multirobot Task Assignment via Consensus ADMM}, 
  year={2023},
  volume={39},
  number={3},
  pages={1781-1800},
  doi={10.1109/TRO.2022.3228132}}

@inproceedings{srikanthan2022resource,
author = {Srikanthan, Anusha and Ravichandar, Harish},
title = {Resource-Aware Adaptation of Heterogeneous Strategies for Coalition Formation},
year = {2022},
isbn = {9781450392136},
publisher = {International Foundation for Autonomous Agents and Multiagent Systems},
address = {Richland, SC},
booktitle = {Proceedings of the 21st International Conference on Autonomous Agents and Multiagent Systems},
pages = {1732–1734},
numpages = {3},
location = {Virtual Event, New Zealand},
series = {AAMAS '22}
}

@article{sucan2012the-open-motion-planning-library,
    Author = {Ioan A. {\c{S}}ucan and Mark Moll and Lydia E. Kavraki},
    Doi = {10.1109/MRA.2012.2205651},
    Journal = {{IEEE} Robotics \& Automation Magazine},
    Month = {December},
    Number = {4},
    Pages = {72--82},
    Title = {The {O}pen {M}otion {P}lanning {L}ibrary},
    Note = {\url{https://ompl.kavrakilab.org}},
    Volume = {19},
    Year = {2012}
}

@article{wu2017gini,
  title={Gini coefficient-based task allocation for multi-robot systems with limited energy resources},
  author={Wu, Danfeng and Zeng, Guangping and Meng, Lingguo and Zhou, Weijian and Li, Linmin},
  journal={IEEE/CAA Journal of Automatica Sinica},
  volume={5},
  number={1},
  pages={155--168},
  year={2017},
  publisher={IEEE}
}

@INPROCEEDINGS{xu2022resource,
  author={Xu, Zirui and Tzoumas, Vasileios},
  booktitle={2022 IEEE 61st Conference on Decision and Control (CDC)}, 
  title={Resource-Aware Distributed Submodular Maximization: A Paradigm for Multi-Robot Decision-Making}, 
  year={2022},
  volume={},
  number={},
  pages={5959-5966}}

@ARTICLE{zhang2025energy,
  author={Zhang, Lixuan and Zhao, Jianzhuang and Lamon, Edoardo and Wang, Yabin and Hong, Xiaopeng},
  journal={IEEE Transactions on Automation Science and Engineering}, 
  title={Energy Efficient Multi-Robot Task Allocation Constrained by Time Window and Precedence}, 
  year={2025},
  volume={22},
  number={},
  pages={18162-18173}}
